\documentclass[entropy,article,accept,moreauthors,pdftex,12pt,a4paper]{mdpi} 
\lastpage{x}
\doinum{10.3390/e16052789}
\pubvolume{16}
\pubyear{2014}
\history{Received: 28 February 2014; in revised form: 28 April 2014  / Accepted: 4 May 2014 / \\Published: 21 May 2014}
\pdfoutput=1

\usepackage{soul}
\usepackage{array, dcolumn, multirow}
\usepackage{hyphenat}
\usepackage{enumitem,caption,booktabs}
\setitemize{parsep=0pt,itemsep=0pt,leftmargin=.4in}
\setenumerate{parsep=0pt,itemsep=0pt,leftmargin=.4in}

\usepackage{subcaption}
\usepackage{tikz}
\usepackage{pgfplots}
\usepackage{amsmath}
\usepackage{amsfonts}
\usepackage{amssymb}
\usepackage{tkz-graph}
\usetikzlibrary{calc,arrows,plotmarks}

\Title{Changing the Environment Based on Empowerment as\linebreak Intrinsic Motivation}

\Author{Christoph Salge *, Cornelius Glackin and Daniel Polani}

\address[1]{%
Adaptive Systems Research Group, University of Hertfordshire, College Lane, \linebreak Hatfield AL10 9AB, UK; E-Mails: c.salge@herts.ac.uk; c.glackin2@herts.ac.uk; d.polani@herts.ac.uk}

\corres{E-Mail: c.salge@herts.ac.uk;\linebreak Tel.:+44-1707-284490.}

\abstract{One aspect of intelligence is the ability to restructure your own environment so that the world you live in becomes more beneficial to you. In this paper we investigate how the information-theoretic measure of agent empowerment can provide a task-independent, intrinsic motivation to restructure the world. We show how changes in embodiment and in the environment change the resulting behaviour of the agent and the artefacts left in the world. For this purpose, we introduce an approximation of the established empowerment formalism based on sparse sampling, which is simpler and significantly faster to compute for deterministic dynamics. Sparse sampling also introduces a degree of randomness into the decision making process, which turns out to beneficial for some cases. We then utilize the measure to generate agent behaviour for different agent embodiments in a Minecraft-inspired three dimensional block world. The paradigmatic results demonstrate that empowerment can be used as a suitable generic intrinsic motivation to not only generate actions in given static environments, as shown in the past, but also to modify existing environmental conditions.\linebreak In doing so, the emerging strategies to modify an agent's environment turn out to be meaningful to the specific agent capabilities, \emph{i.e.}, de facto to its embodiment.}

\keyword{empowerment; intrinsic motivation; information theory}

\begin{document}

\newpage

\section{Introduction}
\vspace{-12pt}

\subsection{Motivation}

One of the remarkable feats of intelligent life is that it restructures the world it lives in for its own benefit. Beavers build dams for shelter and to produce better hunting grounds, bees build hives for shelter and storage, humans have transformed the world in a multitude of ways. Intelligence is not only the ability to produce the right reaction to a randomly changing environment, but it is also about actively influencing changes in the environment, leaving artefacts and structures that provide benefits in the future. In this paper, we want to explore if the framework of \emph{intrinsic motivation} \cite{Ryan200054,Oudeyer2007} can help us to understand and possibly reproduce this phenomenon. In particular, we show some simple, exploratory results on how \emph{empowerment} \cite{klyubin2005empowerment}, as one example of intrinsic motivation, can produce structure in the environment of an agent. 

The basic idea of intrinsic motivation is that an agent's behaviour, or decision making, is not at first guided by some form of externally specified reward, but rather by the optimization of some\linebreak agent-internal measure, which then gives rise to complex and rich interactions with the world. Examples in humans might be curiosity, or the need to be in control of one's environment. Biologically speaking, intrinsic motivation is a proximate cause \cite{scott2011evolutionary} for behaviour, while the ultimate cause is inclusive fitness. Both causes offer an explanation, but of different kinds. To illustrate, the question of why an organism reliably displays a certain behaviour can always be answered with ``evolution'', because ultimately every aspect of an organism has been subject to evolution and is therefore somehow tied to an increase in inclusive fitness. This is the ultimate cause. The proximate cause is more concerned with how evolution causes a certain behaviour or feature. A behavioural drive such as curiosity is like a mechanism that is selected for, because on average, it increases the inclusive fitness of a population. The mechanism of curiosity then causes certain behaviour, as a proximate cause, while the ultimate cause is still evolution, or inclusive fitness. Note that this two-step process also allows for behaviour that is detrimental to survival to ultimately be caused by evolution. A population can select for a certain feature or behaviour, because on average, it increases the fitness, but then circumstances might change, and the previously beneficial behaviour is now detrimental, but still ultimately caused by evolution. Similarly, a mechanism might be mostly beneficial, but might reliably produce detrimental behaviour in certain circumstances.\linebreak This behaviour is proximately caused by the mechanism, but ultimately caused by evolution. 

Different intrinsic motivations, such as the urge to learn, provide proximate explanations why certain behaviour is selected. Ultimately, the idea is that intrinsic motivation offers a generic and universal benefit to an organism, not because it solves a specific problem, such as finding the right food,\linebreak but because it is beneficial for a wide range of scenarios. We are motivated to formalize possible mechanisms for internal motivation in natural agents, so we can then deploy them on artificial agents or robots to generate task-independent but generally sensible behaviour. 

There are several requirements that a good quantitative formulation of intrinsic motivation should meet. Ideally, it should be
\begin{itemize}
\item task-independent,
\item computable from the agent's perspective,
\item directly applicable to many different sensory-motoric configurations, without external tuning
\item sensitive to and reflective of different agent embodiments.
\end{itemize} 

The task-independence demarcates this approach from most classical AI techniques, such as reinforcement learning \cite{sutton-barto:rl:1998}; the general idea is not to solve any particular task well, or to be able to learn how to do a specific task well, but instead to offer an incentive for behaviour even if there is currently no specific task the agent needs to attend to. In a way, intrinsic motivation corresponds to reinforcement learning-based learning strategies as unsupervised learning corresponds to supervised learning.\linebreak It depends, for its success, on the complex and rich interactions of the agent with its environment. The computability from an agent's perspective is an essential requirement. If some form of intrinsic motivation is to be realized by an organism or deployed on an autonomous robot, then the organism/robot needs to be able to evaluate this measure, from its own perspective, \emph{i.e.}, based on its own sensor input. Basically, an agent should be able to evaluate its intrinsic motivations based on its own sensor values and world models, and should not depend on external feedback concerning its performance. Together with the importance of ensuring the response is generic, this is an important constraint on potential candidates for models of intrinsic motivation. In essence, an agent should always have all the information in order to determine its next action. Therefore, if a measure is supposed to be used to determine an agent's next action, then it needs to be computable from the information available to the agent. The applicability to different sensory-motoric configurations derives from our goal to make the model biologically plausible. From an evolutionary perspective, a suitable model must be able to flexibly handle changes in morphological setup of organisms, such as the development of an additional sensor, to ensure that the evolved motivational drives remain viable. Also, a suitable motivational measure would be expected to carry over at least to some extent when the morphological structure of an agent (and its embodiment) changes during an agent's lifetime. For example, if a human or an animal were to lose a leg, or become blind, we would expect that their internal motivations would remain, and adapt to the new situation. The applicability to different sensor-motoric configurations, combined with the requirement of task-independence, basically becomes a requirement for the universality of the measure. More precisely: to be really universal, a measure for intrinsic motivation should ideally operate in essentially the same manner and arise from the same principles, regardless of embodiment or particular situation. That measure can then even be used to identify ``desirable'' changes in situation (e.g., in the context of behaviours) or embodiment (e.g., in the context of development or evolution).

The most relevant property for this paper is the sensitivity of the selected measure of intrinsic motivation to the particular embodiment of an agent. One central idea of enactivism \cite{varela1992embodied} is that the mind can only make sense of the world via continued interaction. As this is mediated through the embodiment of an agent, the body shapes the way an agent thinks \cite{gallagher2005body}. Taking one step further, the idea of morphological computation posits that the body does actually compute, \emph{i.e.}, is an integral part of the thought process, and thereby produces if not the thought, then at least the behaviour of embodied organisms \cite{pfeifer2007body}. If we now look at the whole interaction cycle \cite{von1909umwelt} between environment, sensory perception, action and environment again, then, in full consequence, it is reasonable to assume, that the body not only shapes the mind, but that also, indeed, it is the environment that gets shaped by this mind. 

To illustrate, humans do not just restructure their environment in some random way, but they usually do so to increase the affordances \cite{gibson1979ecological} offered by the environment. As a result, the world humans created reflects their body shapes and available actions and sensors. Imagine an explorer that finds the lost ruins of Machu Picchu \cite{wright2000machu}. Chances are that he is still able to immediately realize that these ruins have been constructed by humans, even if there are no remaining living humans, and that their culture diverged from the explorer's hundreds of years ago. This is possible, because the explorer is looking at stairs that are made for humans to ascend, at doors that fit a human shape, and at shelters that offer what humans need to survive. What we basically aim to do in the present paper, on a much simpler scale, is to connect all these elements, and see what happens if we take empowerment as an intrinsic motivation and apply it to a simple agent with different embodiments that we allow to manipulate and modify its environment.

\subsection{Overview}

The remaining paper begins by first giving a quick overview of different intrinsic motivation approaches, and then introduces empowerment---the intrinsic motivation used in this paper---in more detail. We introduce a simple way to implement approximative estimation technique for empowerment that works for a deterministic and discrete world by sparse sampling. We will then discuss its advantages\linebreak and disadvantages.

Based on this technique, we then present three experiments where an empowerment-controlled agent manipulates a simple three dimensional block world. First, we demonstrate how different embodiments result in the agent producing discernibly different worlds with different characteristics. Then we show how the agent deals with hazards to ensure, first, its survival, and second, that the world becomes more accessible to the agent. Finally, we note that the approximation we use carries an inherent randomness. This leads to different adaptations to the challenges posed by the environment-effectively different ``niches'' that the agent may be able to explore and realize.

\section{Intrinsic Motivation}

\emph{Intrinsic motivation} takes its name from the field of developmental psychology and is defined by Ryan and Deci (\cite{Ryan200054}, p. 56) as: 
\begin{quotation}
Intrinsic motivation is defined as the doing of an activity for its inherent satisfaction rather than for some separable consequence. When intrinsically motivated, a person is moved to act for the fun or challenge entailed rather than because of external products, pressures, or rewards.
\end{quotation} 

Oudeyer and Kaplan \cite{oudeyer2007intrinsic} classify different approaches to intrinsic motivation and argue that it is essential for the development of children, and might, for artificial systems, be a strategy to enable open-ended cognitive development. They also demonstrate how their particular implementation of a curiosity mechanism can push a robot towards situations that optimize its learning rate \cite{Oudeyer2007}. While they acknowledge different approaches to quantify intrinsic motivation, there seems to be a consensus that the possession of such a cognitive ability can be beneficial even if it is not directly tied to a direct external reward and is crucial for the development of complex cognitive interactions with the world. 

Approaches in the field of Artificial Intelligence have also explored the ideas now summarized as intrinsic motivation. An early example is Schmidhuber's work on \emph{Artificial curiosity} \cite{schmidhuber1991curious,schmidhuber2010formal} where an agent receives an internal reward depending on how ``boring'' the environment is that it is currently learning. The most rewarding world for such an agent is one that offers novelty (unexpected states according to the internal model) that can be learned, meaning they are not actually truly random. Another approach is the \emph{autotelic principle} by Steels \cite{steels2004autotelic}, which tries to formalize the concept of\linebreak ``Flow'' \cite{csikszentmihalyi2000beyond}: an agent tries to maintain a state where learning is challenging, but not overwhelming \cite{gordon12:_hierar_curios_loops_activ_sensin}. A common method is to combine the intrinsically motivated reward with classical learning methods, such as the work of Kaplan \cite{kaplan2004maximizing}, which aims to maximise the learning progress of different classical learning approaches by introducing rewards for better predictions of future states.

In general, there are several applications that demonstrate how intrinsic motivation models related to learning (such as curiosity) enhance learning progress and success. More relevant for the current research in this paper are intrinsic motivation approaches that focus on generating behaviour, such as the idea of \emph{homeokinesis}\cite{der1999homeokinesis}, which can be considered as a dynamic version of homoeostasis. With this approach, and its information-theoretic extension of \emph{predictive information} \cite{Ay2008}, it was possible to generate a wide range of robotic behaviours that were dependent on both the environment and the embodiment of the agent \cite{der2012playful}. Importantly, the intrinsic reward can be computed from the agent's perspective, by evaluating how much information the agent's past sensor states possess about the next sensor state. 

The models we \scalebox{.95}[1.0]{presented up to now consider processual models of intrinsic motivation. The motivation} emerges through comparing the learning process or information gleaned in the past with new information coming into the agent. This is a trajectory-oriented view of intrinsic motivation, which orients itself on the life trajectory that the agent or organism has travelled. However, an orthogonal view is possible. One can consider the potential (not actual) trajectories available to an agent, asking what an agent can do now from the state it is currently in. These essentially define the possible futures of the agent and are essentially a property of the environmental dynamics that does not depend on the particular agent history. This is basically a Markovian view, where the agent's decision making does not consider the past but is only based on the current state of the agent. In simpler terms, we can ask, instead of what would be a good thing to do, what would be a good state to be in? This view complements the intrinsic motivations that focus on learning and exploration as it offers incentives for behaviour even if there is nothing new to learn or experience in the moment. In the following, we discuss a particular incarnation of such a model for intrinsic motivation called empowerment. 

\section{Empowerment}

\emph{Empowerment} was introduced \cite{klyubin2005empowerment} to provide an agent with an intrinsic motivation that could act as a stepping stone towards more complex behaviour. It is an information-theoretic measurement that formalizes how much causal influence an agent has on the world it can perceive. Behind its motivation is the idea to unify several seemingly disparate drives of intelligent beings, such as accumulating money, becoming a leader in a gang, staying healthy, maintaining a good internal sugar level, \emph{etc.} \cite{klyubin2008keep}. Ultimately, all these drives enhance survivability in some way, but there seems to be a unifying theme of being in control of one's environment that ties them together. Empowerment attempts to capture this notion in a quantifiable formalism. In this paper, we are particularly interested in the idea of empowerment as a cognitive function producing behaviour \cite{salge2014book}, based on the idea that organisms act so as to improve or maintain their empowerment. 

A number \scalebox{.95}[1.0]{of ideas that would suggest how to unify these aspects stem from psychology. To motivate} \linebreak the empowerment concept, we mention them briefly, before we introduce the concept formally. Oesterreich \cite{oesterreich} argues that agents should act so that their actions lead to perceivably different outcomes, which he calls ``efficiency divergence'' (``Effizienzdivergenz'' in German), so ideally different actions should lead to different perceivable outcomes. Heinz von Foerster famously stated ``I shall act always so as to increase the total number of choices '' \cite{von2003disorder}, arguing that a state where many options are open to an agent is preferable. Furthermore, Seligman \cite{seligman1975helplessness} introduced the concept of ``learned helplessness'', arguing that humans learn to avoid states where one's actions seem to have random or no outcomes, and that being forced to be in such states can have negative consequences for mental health. This relates to more recent empirical studies \cite{trendafilov2013} of humans, which indicated that humans in a control task associate a low level of \emph{empowerment} with frustration, and perform better in situations where they are highly empowered. More recently, ideas similar to empowerment have also emerged in physics \cite{wissner2013causal} proposing a very similar action principle which is however, motivated by the hypothesized thermodynamic Maximum Entropy Production Principle, instead of being based on evolutionary and psychological arguments. We now turn to a formal definition of empowerment that underpins our subsequent studies. Empowerment is formalized as the maximal potential causal flow \citep{Ay2008a} from an agent's actuators to an agent's sensors at a later point in time. This can be formalized in the terms of information theory as Shannon \cite{Shannon1948} channel capacity. To define these terms, consider a discrete random variable $X$ with values $x \in \mathcal{X}$, and the probability of outcome $x$ being $p(x) = Pr \{ X = x \}$. The \emph{entropy} or self-information of $X$ is then defined as
\begin{equation}
\label{entropy}
H(X) = - \sum_{x \in \mathcal{X}} p(x) \log p(x).
\end{equation}
Entropy can be understood as a quantification of uncertainty about the outcome of $X$ before it is observed, or as the average surprise at the observation of $X$. Introducing another random variable $Y$ jointly distributed with $X$ enables the definition of \textit{conditional entropy} as
\begin{equation}
\label{conditionalEntropy}
H(X|Y) = - \sum_{x \in \mathcal{X}} p(y) \sum_{y \in \mathcal{Y}} p(x|y) \log p(x|y).
\end{equation}
This measures the remaining uncertainty about the variable $X$ when a jointly distributed variable $Y$ is known. Since Equation~(\ref{entropy}) is the general uncertainty of $X$, and Equation~(\ref{conditionalEntropy}) is the remaining uncertainty, once $Y$ has been observed, their difference, called \textit{mutual information}, quantifies the average information one can gain about $X$ by observing $Y$. Mutual information, the reduction of uncertainty about one variable when another variable is observed, is defined as
\begin{equation}
\label{mutualInformation}
I(X;Y) = H(X) - H(X|Y).
\end{equation}
The mutual information is symmetric (see \cite{Cover1991}), and it holds that
\begin{equation}
\label{mutualInformationSymmetry}
I(X;Y) = H(X) - H(X|Y) = H(Y) - H(Y|X).
\end{equation}
Finally, a quantity that is used in communication over a noisy channel to determine the maximum information rate that can be reliably transmitted is given by the \emph{channel capacity}:
\begin{equation}
\label{channelCapacity}
C = \max_{p(x)} I(X;Y)\;.
\end{equation}
This maximisation is over all possible input distributions for $p(x)$ where $p(y|x)$, the channel characteristic, is kept fixed.

\subsection{Empowerment in the Perception--Action Loop}
\label{sec:empow-perc-acti}

\begin{figure}[H]
\centering
\begin{tikzpicture}
\GraphInit[vstyle=Dijkstra]
\SetVertexMath \renewcommand*{\VertexLineColor}{white}
\SetGraphUnit{1.5}
\Vertex[L=R_{t+1},Lpos=90]{R0}
\SOEA[L=S_{t+1},Lpos=-90](R0){S0}
\EA[L=A_{t+1},Lpos=-90](S0){A0}
\NOEA[L=R_{t+2},Lpos=-90](A0){R1}
\SOEA[L=S_{t+2},Lpos=-90](R1){S1}
\EA[L=A_{t+2},Lpos=-90](S1){A1}
\NOEA[L=R_{t+3},Lpos=-90](A1){R2}
\WE[empty](R0){E0}
\SOWE[L=A_{t}](R0){E1}
\EA[empty](R2){E2}
\SOEA[L=S_{t+3}](R2){E3}
\SetUpEdge[style={post}]
\tikzstyle{EdgeStyle}=[line width=.8pt]
\tikzstyle{LabelStyle}=[left=3pt]
\tikzstyle{LabelStyle}=[above=3pt]
\Edge(R0)(R1)
\Edge(R1)(R2)
\Edge(R0)(S0)
\Edge(S0)(A0)
\Edge(A0)(R1)
\Edge(R1)(S1)
\Edge(S1)(A1)
\Edge(A1)(R2)
\Edge(R2)(E3)
\Edge(E1)(R0)
\SetUpEdge[style={post}, color=gray]
\tikzstyle{EdgeStyle} = [line width=.8pt]
\Edge(E0)(R0)

\Edge(R2)(E2)

\SetUpEdge[style={post,dotted,color=red,bend right = 20}]
\Edge(E1)(E3)
\Edge(A0)(E3)
\Edge(A1)(E3)
\end{tikzpicture}

\caption{The perception--action loop visualised as a Bayesian network. \textit{S} is the sensor, \textit{A} is the actuator, and \textit{R} represents the rest of the system. The index $t$ indicates the time at which the variable is considered. This model is a minimal model for a simple memoryless agent. The red arrows indicate the direction of the potential causal flow relevant for \protect\linebreak 3-step empowerment.}
\label{fig:pal1}
\end{figure}
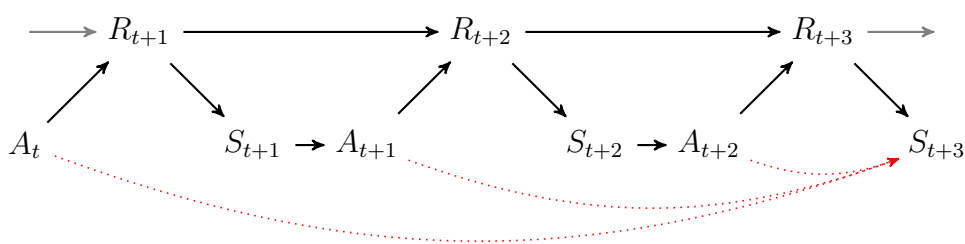

To compute empowerment we model the agent--world interaction as a perception--action loop as seen in Figure~\ref{fig:pal1}. The perception--action loop tells us how actions in the past influence a state in the future. This can be interpreted as a probabilistic channel, where the agent sends actions through a channel and considers the outcome in $S$. The influence of an agent's actions on its future sensor states (not its actual actions, but its \emph{potential} actions) can now be modelled formally as channel capacity: what the agent may (but not necessarily) send over the channel---or in fact, what the agent may (but not necessarily) change in the environment, at the end of its action sequence. To understand this in detail, let us first take a step back and see how to model an agent's interaction with the environment as a causal Bayesian network\linebreak (CBN) \cite{pearl00:_causal}. We define the following three random variables:
\begin{description}
\item[$A$:] the agent's actuator which takes values $a \in \mathcal{A}$. 
\item[$S$:] the agent's sensor, which takes values $s \in \mathcal{S}$
\item[$R$:] the state of the environment, which takes values $r \in \mathcal{R}$
\end{description}
Each of the random variables can consist of several variables of the same type, which is the usual and more general case. For example, saying \emph{actuator} implicitly includes the case of multiple actuators. Multiple actuators (which can be independent of each other) can always be written as being incorporated into one single actuator variable.

The relationship of the random variables can be expressed as a time-unrolled CBN, as seen in Figure~\ref{fig:pal1}.

We are looking at a time-discrete model where an agent interacts with the world. An agent chooses an action $A_{t}$ for the next time step based on its sensor input $S_t$ in the current time step $t$. This influences the state of the world $R_{t+1}$ (in the next time step), which in turn influences the sensor input $S_{t+1}$ of the agent at that time step. The cycle then repeats itself, with the agent choosing another action in $A_{t+1}$. Note that in a more general model this choice of action might also be influenced by some internal state of the agent that carries information about the agent's past.

Empowerment is then defined as the channel capacity between the agent's actuators $A$ and its own sensors $S$ at a later point in time. For example, if we look at the empowerment in regard to the next time step, then empowerment can be expressed as
\begin{equation}
\mathfrak{E} := C(A_t\rightarrow S_{t+1}) \equiv \max_{p(a_t)} I(S_{t+1};A_t)\;.
\label{eq:empowerment}
\end{equation}

Note that the maximization implies that it is calculated under the assumption that the controller choosing the action $A$ is free to act and not bound to a particular behaviour strategy $p(a|s)$. Importantly, the distribution $p^*(a)$ that \emph{achieves} the channel capacity is different from the one that \emph{defines} the actions of an empowerment-driven agent. Empowerment considers only the \emph{potential} information flow, so the agent will only calculate how it \textit{could} affect the world, rather than actually carrying out its potential.

\subsection{$n$-Step Empowerment}

In the simplest version, we considered empowerment as a consequence of a single action taken and the sensor being read out in the subsequent state. However, empowerment, as a measure of the sensorimotor efficiency, may start distinguishing the characteristics of the agent--environment interaction only after several steps have been taken. Therefore, a common generalization of the concept is \textit{$n$-step} empowerment. In this case we consider not a single action variable, but actually a sequence of action variables for the next $n$ time steps: $(A_{t},\dots,A_{t+n-1})$. We will sometimes condense these into a single action variable $A$ for notational convenience. The sensor variable is the resulting sensor state $n$ time steps later, \emph{i.e.}, $S_{t+n}$; again sometimes denoted by $S$. Though it is not the most general treatment possible, here we will consider only ``open loop'' action sequences, \emph{i.e.},\ action sequences that are selected in advance and then carried out without referring to a sensor observation until the final observation $S_{t+n}$. This drastically simplifies both computation and theoretical considerations, as the different possible action sequences $A$ can be treated as if they were separate atomic actions with no inner structure.

\subsection{Model Acquisition}

Looking at the perception--action loop and the formal definition of empowerment, we can see that to compute empowerment, an agent needs a causal model of how its actions in $A$ influence the relevant future states of $S$ in the current state of $R$. There is no need to understand the overall dynamics of the world; it is sufficient for the agent to obtain a model $p(s|a,r)$ of how its actions $a$ influence its future sensors $s$ for the actual states of the world $r$. Therefore, to compute empowerment in a given state $r$, all the agent needs to know are the probability transitions for the states of the world reachable with action sequences $a$, and their causally resulting sensor inputs. It should be noted that local empowerment can reflect global properties. Previous studies of empowerment on small world graphs suggest that the local empowerment reflects global measures of graph centrality for some typical transition graph structures \cite{anthony2008preferred}.

The acquisition of a local causal model can be done in various ways. For instance, previous work in the continuous domain demonstrated how Gaussian Process learners can learn the local dynamics of a system, and empowerment based on these models can then be used to control pendulums \cite{jung2011empowerment} or even a physical robot \cite{leu2013}. However, note that the quality of the model can influence the quality of the empowerment calculation \cite{Salge2012}. This becomes particularly interesting if the environment offers a localized source of noise \cite{salge2013empowerment}, which would induce an empowerment-driven agent to avoid certain states since the agent predicts that its empowerment would suffer in these. Nevertheless, it could also be the case that such specific states are not a true source of noise, but that the agent has just not explored this part of the state space enough to model the relevant probability transitions sufficiently, and therefore just assumes that a certain state leads to a noisy outcome. In this case, it would be better for the agent's long term empowerment to explore the state in more detail and update its internal model. 

The effect of learning and incomplete models on empowerment is a promising avenue for further research; however, apart from qualitative insights (such as the above) not much is known about the theory. A detailed theoretical study has not been undertaken yet and our understanding of its ramifications must be considered to be only in its beginning. At the present stage, and for the present purpose, we will utilize the fact that model acquisition and empowerment computation can be completely separated in our case. As we are more interested in the \textit{effects} of empowerment, in the following we will further assume that a suitable and exact causal model has been acquired in some fashion, and that this model will be the basis for our computation of empowerment. 

\subsection{Sensor Limitation}

While empowerment is defined as the channel capacity from an agent's actuators to an agent's sensors, early work \cite{klyubin2005empowerment} equated $S$ and $R$, basically assuming that the agent could see the whole world state.\linebreak In this case, empowerment becomes a measure of causal influence on the overall world, rather than on the world the agent can \emph{perceive}. In general, the agent's sensor state $S$ is a function of the state of the world $R$, with possibly some added noise. Limiting the sensor to contain only part of the information of $R$ has two different effects on empowerment.

Evaluating empowerment from the agent's perspective requires the agent to determine the current state of $R_t$; the current state of the environment. If the sensor is limited, so the agent can only determine the state of $R_t$ with some degree of certainty, then the agent-estimated empowerment will be\linebreak lower \cite{klyubin2008keep}. The worst case is the one where the agent has no idea what state the world is in, which is called \emph{context-free} empowerment. This can be illustrated by a simple grid world example. Imagine the agent lives in an infinite two-dimensional grid world, with coordinates $x$ and $y$. It can move up, down, left and right. If the agent knows that it is currently at coordinate $(0,0)$, then the conditional probabilities $p(s|a)$ are clear, and we can easily compute the \emph{context-dependent} empowerment. Moving up, for example, will take the agent to coordinate $(1,0)$. Assuming there is no noise, the empowerment then computes to 2 bit, as the 4 different actions reliably lead to 4 different, predictable outcomes. If, however, the agent does not know where it currently is, then the outcome of its action, \emph{i.e.}, the resulting sensor input, remains completely uncertain, and there is no context-free empowerment.

In general, context-\scalebox{.95}[1.0]{free empowerment is always lower or equal than context-dependent empowerment,} so having any kind of sensor information about the starting state of the world can only increase empowerment. Having perfect world information, which is basically like computing the empowerment for the agent from \scalebox{.95}[1.0]{an outside perspective, results in the highest possible context-dependent empowerment.} However, it is interesting to note that the highest possible context-dependent empowerment can already be achieved with a sensor that can differentiate all those states with different probability distributions $p(s|a)$. Hence, all states where all different actions $a \in A$ lead to the same probability distributions $p(s)$ can be lumped together without loss of subjective empowerment. This is similar to the idea of epsilon machines \cite{shalizi2001}, and has been used to define an empowerment-based context for an agent by basically looking for those partitions of the sensor that have the highest empowerment \cite{klyubin2008keep}.\linebreak In our examples in this paper, we assume that the agent knows exactly what state $R_t$ the world is in. 

So far, we looked at the effect of sensor limitations of $S_t$ in regard to $R_t$, the ``starting'' state of the environment for which empowerment is evaluated. Sensor limitations also affect empowerment if the sensor state $S_{t+n}$ resulting from the agent's actions (after $n$ steps) only reflects part of the resulting world state $R_{t+n}$. This can be either caused by an actual sensor limitation or by the agent choosing to only consider part of its resulting sensor input for the computation of empowerment. This has been briefly discussed in \cite{salge2014book}, where we argue that the selection of certain parts of the sensor input for the empowerment computation can (intentionally) focus empowerment on specific aspects of the world. For example, restricting the empowerment computation to location sensors makes empowerment concerned with being in a state where the agent has the greatest potential causal influence on its future location. Also, there might be aspects of the world that the agent cannot influence in either case, which would mean they could safely be ignored without changing the empowerment computation. Following this idea, empowerment has been applied to sensor evolution \cite{klyubin2008keep}, which showed that some parts of the resulting sensor states are less relevant for empowerment than other parts. Given an additional cost for each sensor, certain sensors would reliably disappear, while others provided enough empowerment to justify their cost. 

From an evolutionary perspective, empowerment, as a model of effective action, can be seen as an informational ``impedance match'' of the available actions of the agent and its sensors. It would not make sense to sustain actuator potentialities on an evolutionary scale whose effect would never materialize in the available sensors in the niche in which an organism operates; they could be evolved away without loss of performance. Vice versa, if one finds that these choices evolved, it would indicate that there are scenarios where they provide an evolutionary advantage.

As a corollary, using empowerment maximisation to generate behaviour for an agent who has actually evolved its sensors (and only has a limited access to the overall world state $R$) would drive an agent/organism towards states where its actuators are most effective, as this best matches the niche for which the agent has evolved (for a more detailed discussion of this point, see also \cite{klyubin2008keep}).

In this paper, we will also only consider parts of the agent's resulting sensors for empowerment.\linebreak Thus, while the agent has full access to its sensors to determine its current state (\emph{i.e.}, context), it will only be concerned with the potential causal effect of its action sequence at the end location. 

%
\newpage

\subsection{Applicability of Empowerment}

Initially, most empowerment work focused on systems with discrete action and sensor\linebreak variables \cite{klyubin2005empowerment,klyubin2008keep}. If there is no noise in the channel, then empowerment, \emph{i.e.}, the maximal mutual information between actions and sensors, simply reduces to the logarithm of the number of different resulting sensor states for all actions. If there is noise in the channel, then classical channel capacity algorithms, such as the Blahut--Arimoto algorithm \cite{blahut1972computation,arimoto1972}, can be used to determine the channel capacity. However, this quickly becomes infeasible for larger horizons $n$ in $n$-step empowerment, because the number of action sequences to consider grows exponentially with $n$. 

We would like to point out that empowerment itself can be extended to more generic domains.\linebreak The idea of impoverished empowerment \cite{anthony2011impoverished,anthony13:_gener_self_motiv_strat_ident} deals with large numbers of action sequences by only considering a selected subset of sequences that contribute most to the empowerment of the agent, treating them as meta-actions to build longer action sequences. Also, empowerment is well-defined for continuous action and sensor variables, but there is no general applicable algorithm to compute the channel capacity for general continuous variables. Jung proposes an approximation via Monte-Carlo sampling \cite{jung2011empowerment}, and there is an even faster approximation based on the idea of modelling the channel as a linear Gaussian channel \cite{telatar1999capacity,salge2013empowerment,salge2014book}.

In this paper we will focus solely on deterministic and discrete empowerment, and we will introduce a new sparse sampling technique to estimate empowerment. Our new technique offers some new features such as the inclusion of randomness in the decision process, but the main advantage is its fast computation. 

\subsection{Approximation with Sparse Sampling}
\label{sec:sparse}

Empowerment for a deterministic system with discrete action and sensor variables is given by the logarithm of all reachable sensor states, given all available actions. This basically reduces empowerment to a mobility or reachability measure in deterministic worlds. Nevertheless, if the amount of considered action sequences is growing large, this can be hard to compute, since every single action sequence needs to be checked to determine if it reaches a new sensor state. However, there is a special configuration: if the number of reachable sensor states is small compared with the number of action sequences, it might be feasible to approximate the empowerment value by sampling only a subset of all action sequences. Recent advances in Monte-Carlo-based tree-search \cite{browne2012survey} have demonstrated the power of approaches that only sample a subset of the available action sequences. In this subsection, we will theoretically evaluate how good such an approximation will be.

We want to compute empowerment from an agent's actions $a \in A$ to an agent's sensor states \linebreak$s \in S$. Note that if we are dealing with $n$-step empowerment, then $a$ will be one of the action sequences containing $n$ atomic actions. We will use the terms action and action sequences interchangeably, as they can both be treated the same way in regard to their empowerment. 

The system is deterministic, so every action $a$ will result in a specific sensor state $s$. Assume that the state space of the actions is much larger than the state space of the sensors, $|A| \gg |S|$. We can then marginalize the sensor state distribution $p(s)$, under the assumption that the actions have an equal distribution $p(a) = \frac{1}{|A|}$, as $p(s) = \sum_{a \in A} p(s|a)p(a)$. This distribution $p(s)$ is then also the probability to get the resulting state $s \in S$, if a random action $a \in A$ is picked.

Now, importantly: $p(a)$ will almost never be an empowerment-achieving action distribution $p^*(a)$. The latter is characterized (in the deterministic case) by the corresponding outcome distribution $p^*(s)$ being an equidistribution over all reachable states, which $p(s)$ will almost never be. However, $p(s)$ will serve as a proxy to estimate which states are reachable. Once we know that, $p^*(s)$ is trivially known, and the resulting empowerment is $\log(|S^*|)$ where $S^* \subseteq S$ is the set of states that can be reached from the initial position by action sequences of length $n$.

Hence, basically, all $s \in S$ with non-vanishing probabilities $p(s) > 0$ are in $\log(|S^*|)$, and the ``true'' empowerment for this system is $\log(|S^*|)$. The question now is how well will we approximate this ``true'' empowerment value if we only sample $m$ actions. 

\subsubsection{Experimental Evaluation}

First, we evaluate this question quantitatively by running several computer simulations with different distributions $p(s)$ to determine how the sample size $m$ affected the empowerment approximation. Instead of generating action sequences, we just assume, since $|A| \gg |S|$, that if we pick a random action $a$,\linebreak we will observe a resulting state $s$ with a probability of $p(s)$. We considered the following distribution for $p(s)$:
\begin{eqnarray}
p_1(s) &=& (\frac{1}{10},\frac{1}{10},\frac{1}{10},\frac{1}{10},\frac{1}{10},\frac{1}{10},\frac{1}{10},\frac{1}{10},\frac{1}{10},\frac{1}{10})\\
p_2(s) &=& (\frac{91}{100},\frac{1}{100},\frac{1}{100},\frac{1}{100},\frac{1}{100},\frac{1}{100},\frac{1}{100},\frac{1}{100},\frac{1}{100},\frac{1}{100},)\\
p_3(s) &=& (\frac{991}{1000},\frac{1}{1000},\frac{1}{1000},\frac{1}{1000},\frac{1}{1000},\frac{1}{1000},\frac{1}{1000},\frac{1}{1000},\frac{1}{1000},\frac{1}{1000},)\\
p_4(s) &=& (\frac{1}{2},\frac{1}{4},\frac{1}{8},\frac{1}{16},\frac{1}{32},\frac{1}{64},\frac{1}{128},\frac{1}{256},\frac{1}{512},\frac{1}{512})\\
p_5(s) &=& (\frac{1}{5},\frac{1}{5},\frac{1}{5},\frac{1}{5},\frac{1}{10},\frac{1}{50},\frac{1}{50},\frac{1}{50},\frac{1}{50},\frac{1}{50})
\end{eqnarray}
To evaluate the approximation for a given amount of samples, we took $m$ samples according to this distribution and counted how many sensor states $s$ were sampled at least once. Since the model is deterministic, we only need to reach a state once to make sure that we can reliably reach it. The logarithm of this number was then the estimated empowerment for that sampling. For each sample size $m$, we repeated this process 1000 times and averaged the results, resulting in our value of $\mathfrak{E}_m(p(s))$, the estimation of the empowerment for $p(s)$ with $m$ samples.

\begin{figure}[H]
\centering
\begin{tikzpicture}
\begin{axis}[
xlabel= $m$: Number of Samples,
ylabel = $\mathfrak{E}_m$: Estimated Empowerment ,
width = 10cm,
xmax = 1000,
ymin = 0,
grid = both,
legend pos= south east
]
\addplot[no marks] file{Prob1.txt};
\addlegendentry{$p_1(s)$};
\addplot[no marks,blue] file{Prob2.txt};
\addlegendentry{$p_2(s)$};
\addplot[no marks,red] file{Prob3.txt};
\addlegendentry{$p_3(s)$};
\addplot[no marks,green] file{Prob4.txt};
\addlegendentry{$p_4(s)$};
\addplot[no marks,orange] file{Prob5.txt};
\addlegendentry{$p_5(s)$};
\end{axis}
\end{tikzpicture}
\caption{The estimated empowerment values for different amounts of samples, obtained with 1000 different random samplings for each sample number. The different curves are the result of different distributions for $p(s)$. The estimated empowerment value is in nats, the true empowerment value is $\ln(10)$, and the asymptotes of the graphs reach this level.}
\label{approximation}
\end{figure}
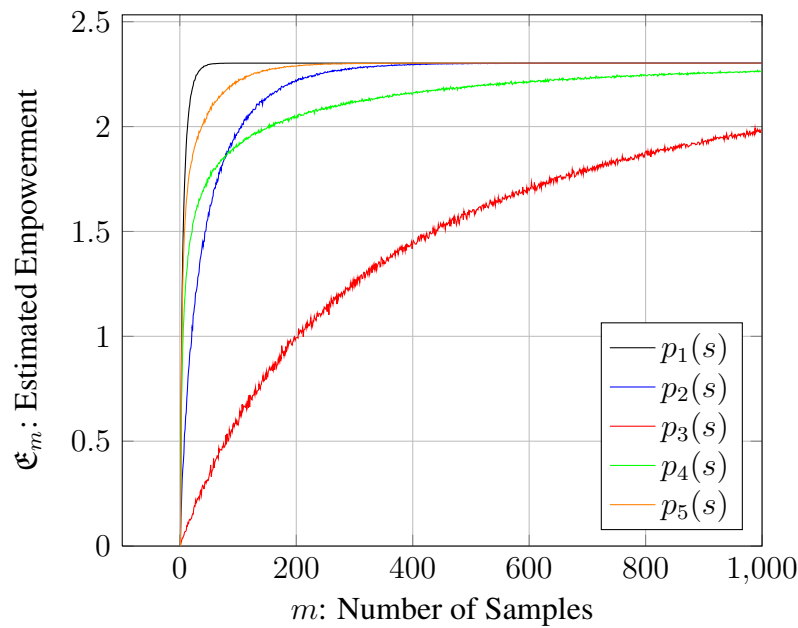

We can see in Figure~\ref{approximation} how the approximation approaches the actual empowerment value of $\ln(10)$ from below as the amount of samples increases. Depending on the distribution of $p(s)$, the amount of samples needed to obtain a good estimate of the actual empowerment varies. The equidistribution $p_1(s)$ already provides a good approximation for a low amount of samples (well below 50), whereas more uneven distributions require a higher amount of samples to reach a similar approximation level. Interestingly, when we compare $p_4(s)$ and $p_2(s)$, we can see that $p_4$ provides a better approximation than $p_2$ for less than 100 samples, while $p_2$ approximates better for more than 100 samples.

\subsubsection{Mathematical Model of Approximation Quality}

The graph in Figure~\ref{approximation} is basically a combination of the different probabilities to discover the different states of $S$ with the sampled action sequences. The probability that a specific state $s \in S$ is found is $1-(1-p(s))^m$, assuming that we have so many actions compared with sensor states that a previously taken action does not change the distribution of resulting states in $p(s)$ for another random action $a$. Alternatively, we can also assume that our sampling process does not remember what actions were taken beforehand but only flags the reached states in $S$. In any case, we can approximate the expected value of how many states we are going to discover as 
\begin{equation}
\sum_{s \in S} 1 - (1-p(s))^m = |S| - \sum_{s \in S}(1-p(s))^m.
\end{equation}
This approximation neglects the fact that there is a dependency between the probabilities for discovering different states $s$, since one sample cannot discover more than one state. Thus, when we only have a low amount of samples, say one, this approximation assumes that there is a chance that each state could be discovered, which simply cannot happen. For example, if we only sample once, then we cannot discover more than one state. This problem becomes negligible for larger samples sizes.

We can now make the empowerment approximation by taking the logarithm of this estimate, so our model to compute the approximation quality given a limited sample size $m$ of random samples is:
\begin{equation}
\mathfrak{E}_m(p(s)) \approx 
\mathfrak{E^*}_m(p(s)) =\log \left( |S| - \sum_{s \in S}(1-p(s))^n\right).
\end{equation}

\subsubsection{Discussion of Sparse Sampling}

\begin{figure}[H]
\centering
\begin{tikzpicture}
\begin{axis}[
xlabel= $m$: Number of Samples,
ylabel = Estimated Empowerment ,
width = 10cm,
xmax = 1000,
ymin = 0,
grid = both,
legend pos= south east
]
\addplot[only marks,mark =x,mark size = 1pt,black] file{Prob1.txt};
\addlegendentry{$\mathfrak{E}_m(p_1(s))$};
\addplot[no marks,black] file{Approx1.txt};
\addlegendentry{$\mathfrak{E^*}_m(p_1(s))$};

\addplot[only marks,mark =*,mark size = 0.2pt,blue] file{Prob2.txt};
\addlegendentry{$\mathfrak{E}_m(p_2(s))$};
\addplot[no marks,blue] file{Approx2.txt};
\addlegendentry{$\mathfrak{E^*}_m(p_2(s))$};

\addplot[only marks,mark =*,mark size = 0.2pt,red] file{Prob3.txt};
\addlegendentry{$\mathfrak{E}_m(p_3(s))$};
\addplot[no marks,red] file{Approx3.txt};
\addlegendentry{$\mathfrak{E^*}_m(p_3(s))$};

\addplot[only marks,mark =*,mark size = 0.2pt,green] file{Prob4.txt};
\addlegendentry{$\mathfrak{E}_m(p_4(s))$};
\addplot[no marks,green] file{Approx4.txt};
\addlegendentry{$\mathfrak{E^*}_m(p_4(s))$};

\addplot[only marks,mark =*,mark size = 0.2pt,orange] file{Prob5.txt};
\addlegendentry{$\mathfrak{E}_m(p_5(s))$};
\addplot[no marks,orange] file{Approx5.txt};
\addlegendentry{$\mathfrak{E^*}_m(p_5(s))$};
\end{axis}
\end{tikzpicture}
\caption{A comparison of the mathematical approximation of approximation quality with the results from the simulation. $p_1$ behaves very similar to the mathematical model. For the other distributions the approximation seems to capture the general trend quite closely, even if the fit is not precise, indicating the usefulness of the model.}
\label{Comparison}
\end{figure}
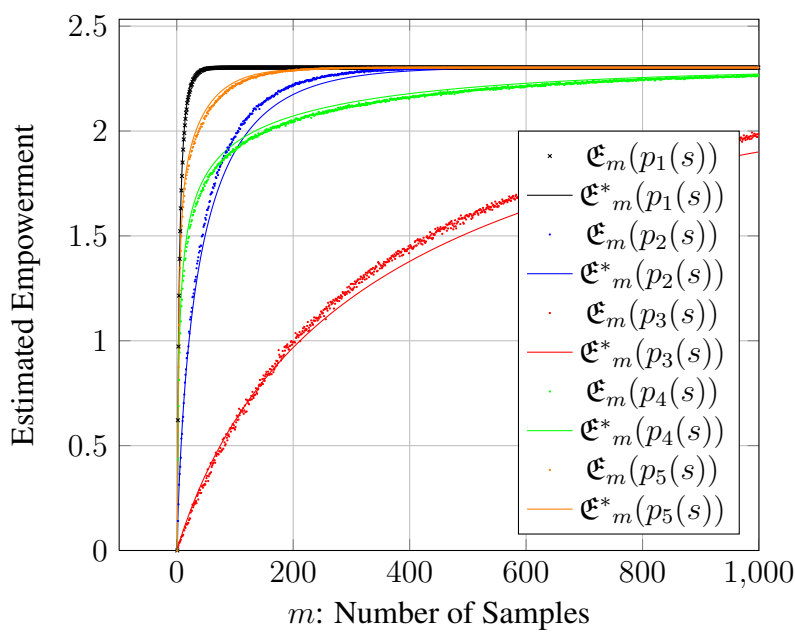

We can see in Figure~\ref{Comparison} that the mathematical approximation gives us a good idea of how well sampling approximates the actual empowerment value. Keep in mind that the actual empowerment for all distributions is the same; it is $\log(10)$. However, the estimated value varies depending on the number of samples and how ``skewed'' the distribution is. Basically, the harder it is to discover rare outcomes, the more empowerment is underestimated. Thus, with sparse sampling, empowerment estimates will be higher (and more accurate) for distributions where the outcomes are easier to discover, \emph{i.e.}, have more action sequences leading to them.

In the experiments in this paper we will use sparse sampling, mainly because the exploding amount of possible actions $a$ for longer $n$-step sequences does not allow for an exhaustive exploration. However, the evaluating of the approximation points at two unexpected advantages that are helpful for our agent behaviour generation.

First, the random selection of sampled sequences introduces a degree of randomness that helps the agent to escape from local minima. Sometimes, all actions will lead to states with lower empowerment than the current state, forcing the agent to stay idle, even if there might be a much higher empowered state just a few steps in one direction or another. The noise incidentally produced by random sampling introduces both a higher degree of exploration and variability in the behaviour \linebreak of the agent.

Secondly---also a cognitively interesting phenomenon---it makes the agent prefer states with ``easier'' empowerment, {\em i.e.}, those states where the resulting states are more evenly distributed between the actions, which usually means states that are reachable with several different action sequences. Arguably, this makes it harder for the agent to lose access to this state, because even if one action sequence leading there is blocked, it may still use another action sequence to reach this sensor state. Similarly, this also gives the agent a somewhat variable time horizon. If there are reachable states that are at the very horizon of the agent's actions, and are only reachable with very few and specific action sequences, then these states might not be discovered at all. If some disturbance would now remove access to these far away states, the agent might not even notice, since it is unlikely that the agent would discover these state in any case. On the other hand, if a much closer state, reachable with several action sequences, becomes blocked, then the agent is more likely to notice this loss and take action accordingly. As a result, sparse sampling empowerment might be more reactive to changes and problems that are closer to the agent in time and space. We note that these two consequences were not designed into the sparse sampling methodology, but turn out to be plausible and attractive features\linebreak of the model. 

\section{Experimental Section}
\vspace{-12pt}

\subsection{General Model}

In this section we now apply control driven by empowerment estimated by sparse sampling to an agent in a three dimensional grid world; a model somewhat inspired by the computer game \linebreak``Minecraft'' \cite{minecraft}. Each location is a block, identified by three integer coordinates. Each location can either be \textit{empty}, filled with \textit{earth}, or filled with an \textit{agent}.

The model progresses in turn; at each turn the agent can take one of twelve different actions, and then the world is updated. There are different types of actions, including moving, interacting, \emph{etc.}, and the agent chooses exactly one action every turn. The standard agent can decide to move in one of the four cardinal directions, called \textit{north}, \textit{east}, \textit{south} and \textit{west}. If the target location is empty, the agent will enter it. If the target location is filled, then there are two options. If the location above the target location is empty, the agent will move in its desired direction and up by one unit (hence a 1-step climb). If the location above is also filled, then the agent's move is blocked, and the agent will not move.

The agent can also decide to interact with the 6 adjacent target locations (\textit{up, down, north, south, east, west}). The interpretation of the interaction is context-dependent on the current state of the agent's inventory, which has space for exactly one block. If the agent's inventory is empty, and the adjacent target location contains something, then the agent will take the block, filling its inventory with it. The location in the world will then be empty. If the agent's inventory is full, then the agent will try to place the block in the target location in the world, succeeding if the relevant adjacent target location is empty. The agent can also \textit{do nothing}, or \textit{destroy} the block in its inventory, resulting in an empty inventory. 

During the world update, the model checks if the agent is above an empty location, in which case the agent will fall down one block (per turn). Earth blocks are not affected by gravity, so it is possible to have ``levitating'' structures. 

The agent chooses its action by maximising the expected sparse empowerment of the state resulting from its action. Every time the agent can take an action, it evaluates all its actions by estimating the empowerment for the state resulting from the said actions. It picks the action that leads to the state with the highest estimated empowerment. In case of multiple candidates, it picks one of the optimal actions\linebreak at random.

Empowerment for a specific state of the world is computed by using the actual world model (so the agent knows exactly how the world will develop). Starting from a specific world state, including the full configuration of all blocks, the agent then evaluates 1000 randomly generated 15-step action sequences, and counts how many different resulting world states there are. To simplify, the resulting world states are only compared in terms of agent location ($x$, $y$ and $z$ coordinate). The estimated empowerment is the logarithm of the different reachable locations. Every time the agent can take an action, it evaluates all its actions by estimating the empowerment for the state resulting from the said actions. It picks the action that leads to the state with the highest estimated empowerment. In case of multiple candidates, it picks one of the optimal actions at random. 
\subsection{Experiment 1: Changing Embodiments}

In our first experiment we want to demonstrate how empowerment, as an intrinsic motivation, is sensitive to the agent embodiment, and how different embodiments are reflected in the changes made to the world. For this, we defined three different types of agents:
\begin{description}
\item[Climbing Agent] This is the standard agent; it can take and place blocks, move in four directions, and will climb up one block if the target block is filled but the one above is empty.
\item[Non-Climbing Agent] Differs from the standard agent in that it will not move up if the target block is filled. Its movement will simply be blocked, just as if there were two blocks on top of each other. 
\item[Flying Agent] Differs from the standard agent in that it can decide to also move up and down, and it is not affected by gravity (does not fall). Its movement will be blocked in any direction if it encounters a block in the target location. 
\end{description}

All three agents are controlled by the same controller, which tries to greedily maximise the 15 step empowerment estimation gained from 1000 samples. All three agents were placed in a 3 $\times$ 3 $\times$ 8 sized world, where the lowest 5 levels where filled with earth blocks. The agents start in the middle of layer 6, just above the ground, as seen in Figure~\ref{fig:nonClimb}.

\newcommand\xaxis{190}
\newcommand\yaxis{-45}
\newcommand\zaxis{90}

\newcommand\topside[4]{
\fill[fill=black!#4, draw=black,shift={(\xaxis:#1)},shift={(\yaxis:#2*0.7)}, shift={(\zaxis:#3)}] (0,0) -- (10:1) -- ++(135:0.7) --++(190:1)--(0,0);
}

\newcommand\leftside[4]{
\fill[fill=black!#4, draw=black,shift={(\xaxis:#1)},shift={(\yaxis:#2*0.7)}, shift={(\zaxis:#3)}] (0,0) -- (0,-1) -- ++(135:0.7) -- ++(0,1)--(0,0);
}

\newcommand\rightside[4]{
\fill[fill=black!#4, draw=black,shift={(\xaxis:#1)},shift={(\yaxis:#2*0.7)}, shift={(\zaxis:#3)}] (0,0) -- (10:1) -- ++(0,-1) --(0,-1)--(0,0);
}

\newcommand\cube[6]{
\topside{#1}{#2}{#3}{#4} \leftside{#1}{#2}{#3}{#5} \rightside{#1}{#2}{#3}{#6}
}

\newcommand\topsidea[3]{
\fill[fill=blue!#30, draw=black,shift={(\xaxis:#1)},shift={(\yaxis:#2*0.7)}, shift={(\zaxis:#3)}] (0,0) -- (10:1) -- ++(135:0.7) --++(190:1)--(0,0);
}

\newcommand\leftsidea[3]{
\fill[fill=blue!60, draw=black,shift={(\xaxis:#1)},shift={(\yaxis:#2*0.7)}, shift={(\zaxis:#3)}] (0,0) -- (0,-1) -- ++(135:0.7) -- ++(0,1)--(0,0);
}

\newcommand\rightsidea[3]{
\fill[fill=blue!80, draw=black,shift={(\xaxis:#1)},shift={(\yaxis:#2*0.7)}, shift={(\zaxis:#3)}] (0,0) -- (10:1) -- ++(0,-1) --(0,-1)--(0,0);
}

\newcommand\cubeAgent[3]{
\topsidea{#1}{#2}{#3} \leftsidea{#1}{#2}{#3} \rightsidea{#1}{#2}{#3}
}

\newcommand\topsidel[3]{
\fill[fill=red!80, draw=black,shift={(\xaxis:#1)},shift={(\yaxis:#2*0.7)}, shift={(\zaxis:#3)}] (0,0) -- (10:1) -- ++(135:0.7) --++(190:1)--(0,0);
}

\newcommand\leftsidel[3]{
\fill[fill=red!90, draw=black,shift={(\xaxis:#1)},shift={(\yaxis:#2*0.7)}, shift={(\zaxis:#3)}] (0,0) -- (0,-1) -- ++(135:0.7) -- ++(0,1)--(0,0);
}

\newcommand\rightsidel[3]{
\fill[fill=red!95, draw=black,shift={(\xaxis:#1)},shift={(\yaxis:#2*0.7)}, shift={(\zaxis:#3)}] (0,0) -- (10:1) -- ++(0,-1) --(0,-1)--(0,0);
}

\newcommand\cubeLava[3]{
\topsidel{#1}{#2}{#3} \leftsidel{#1}{#2}{#3} \rightsidel{#1}{#2}{#3}
}

\begin{figure}[H]
\centering
\begin{subfigure}[b]{0.3\textwidth}
\begin{tikzpicture}[scale=0.6]
\cube{0}{0}{0}{20}{50}{70} 
\cube{0}{1}{0}{20}{50}{70} 
\cube{0}{2}{0}{20}{50}{70} 
\cube{1}{0}{0}{20}{50}{70} 
\cube{1}{1}{0}{20}{50}{70} 
\cube{1}{2}{0}{20}{50}{70} 
\cube{2}{0}{0}{20}{50}{70} 
\cube{2}{1}{0}{20}{50}{70} 
\cube{2}{2}{0}{20}{50}{70} 
\cube{0}{0}{1}{23}{53}{73} 
\cube{0}{1}{1}{23}{53}{73} 
\cube{0}{2}{1}{23}{53}{73} 
\cube{1}{0}{1}{23}{53}{73} 
\cube{1}{1}{1}{23}{53}{73} 
\cube{1}{2}{1}{23}{53}{73} 
\cube{2}{0}{1}{23}{53}{73} 
\cube{2}{1}{1}{23}{53}{73} 
\cube{2}{2}{1}{23}{53}{73} 
\cube{0}{0}{2}{26}{56}{76} 
\cube{0}{2}{2}{26}{56}{76} 
\cube{1}{0}{2}{26}{56}{76} 
\cube{2}{0}{2}{26}{56}{76} 
\cube{2}{1}{2}{26}{56}{76} 
\cube{2}{2}{2}{26}{56}{76} 
\cube{0}{0}{3}{29}{59}{79} 
\cube{0}{1}{3}{29}{59}{79} 
\cube{0}{2}{3}{29}{59}{79} 
\cube{1}{1}{3}{29}{59}{79} 
\cube{2}{2}{3}{29}{59}{79} 
\cube{1}{2}{4}{32}{62}{82} 
\cube{1}{1}{5}{35}{65}{85} 
\cube{2}{1}{5}{35}{65}{85} 
\cube{0}{1}{6}{38}{68}{88} 
\cubeAgent{1}{1}{6} 
\end{tikzpicture}
\caption{}
\label{fig:oneStart}
\end{subfigure}
\begin{subfigure}[b]{0.3\textwidth}
\begin{tikzpicture}[scale=0.6]
\cube{0}{0}{0}{20}{50}{70} 
\cube{0}{1}{0}{20}{50}{70} 
\cube{0}{2}{0}{20}{50}{70} 
\cube{1}{0}{0}{20}{50}{70} 
\cube{1}{1}{0}{20}{50}{70} 
\cube{1}{2}{0}{20}{50}{70} 
\cube{2}{0}{0}{20}{50}{70} 
\cube{2}{1}{0}{20}{50}{70} 
\cube{2}{2}{0}{20}{50}{70} 
\cube{0}{0}{1}{23}{53}{73} 
\cube{0}{1}{1}{23}{53}{73} 
\cube{0}{2}{1}{23}{53}{73} 
\cube{1}{0}{1}{23}{53}{73} 
\cube{1}{1}{1}{23}{53}{73} 
\cube{1}{2}{1}{23}{53}{73} 
\cube{2}{0}{1}{23}{53}{73} 
\cube{2}{1}{1}{23}{53}{73} 
\cube{2}{2}{1}{23}{53}{73} 
\cube{0}{0}{2}{26}{56}{76} 
\cube{0}{1}{2}{26}{56}{76} 
\cube{0}{2}{2}{26}{56}{76} 
\cube{1}{0}{2}{26}{56}{76} 
\cube{1}{1}{2}{26}{56}{76} 
\cube{1}{2}{2}{26}{56}{76} 
\cube{2}{0}{2}{26}{56}{76} 
\cube{2}{1}{2}{26}{56}{76} 
\cube{2}{2}{2}{26}{56}{76} 
\cube{0}{0}{3}{29}{59}{79} 
\cube{0}{1}{3}{29}{59}{79} 
\cube{0}{2}{3}{29}{59}{79} 
\cube{1}{0}{3}{29}{59}{79} 
\cube{1}{1}{3}{29}{59}{79} 
\cube{1}{2}{3}{29}{59}{79} 
\cube{2}{0}{3}{29}{59}{79} 
\cube{2}{1}{3}{29}{59}{79} 
\cube{2}{2}{3}{29}{59}{79} 
\cube{0}{0}{4}{32}{62}{82} 
\cube{0}{1}{4}{32}{62}{82} 
\cube{0}{2}{4}{32}{62}{82} 
\cube{1}{0}{4}{32}{62}{82} 
\cube{1}{1}{4}{32}{62}{82} 
\cube{1}{2}{4}{32}{62}{82} 
\cube{2}{0}{4}{32}{62}{82} 
\cube{2}{1}{4}{32}{62}{82} 
\cube{2}{2}{4}{32}{62}{82} 
\cubeAgent{1}{1}{5} 
\end{tikzpicture}
\caption{}
\label{fig:nonClimb}
\end{subfigure}
\begin{subfigure}[b]{0.3\textwidth}
\begin{tikzpicture}[scale=0.6]
\cube{0}{0}{0}{20}{50}{70} 
\cube{0}{1}{0}{20}{50}{70} 
\cube{0}{2}{0}{20}{50}{70} 
\cube{1}{0}{0}{20}{50}{70} 
\cube{1}{1}{0}{20}{50}{70} 
\cube{1}{2}{0}{20}{50}{70} 
\cube{2}{0}{0}{20}{50}{70} 
\cube{2}{1}{0}{20}{50}{70} 
\cube{2}{2}{0}{20}{50}{70} 
\cube{0}{0}{1}{23}{53}{73} 
\cube{0}{1}{1}{23}{53}{73} 
\cube{0}{2}{1}{23}{53}{73} 
\cube{1}{0}{1}{23}{53}{73} 
\cube{1}{1}{1}{23}{53}{73} 
\cube{1}{2}{1}{23}{53}{73} 
\cube{2}{0}{1}{23}{53}{73} 
\cube{2}{1}{1}{23}{53}{73} 
\cube{2}{2}{1}{23}{53}{73} 
\cube{0}{0}{2}{26}{56}{76} 
\cube{0}{1}{2}{26}{56}{76} 
\cube{0}{2}{2}{26}{56}{76} 
\cube{1}{2}{2}{26}{56}{76} 
\cube{2}{0}{2}{26}{56}{76} 
\cube{2}{1}{2}{26}{56}{76} 
\cube{2}{2}{2}{26}{56}{76} 
\cube{0}{0}{3}{29}{59}{79} 
\cube{0}{2}{3}{29}{59}{79} 
\cubeAgent{1}{1}{3} 
\cube{2}{0}{3}{29}{59}{79} 
\cube{2}{2}{3}{29}{59}{79} 
\cube{0}{0}{4}{32}{62}{82} 
\cube{0}{2}{4}{32}{62}{82} 
\cube{2}{0}{4}{32}{62}{82} 
\cube{2}{2}{4}{32}{62}{82} 
\end{tikzpicture}
\caption{}
\label{fig:fly}
\end{subfigure}
\caption{The representative resulting worlds for (\textbf{a}) a climbing, (\textbf{b}) a non-climbing and (\textbf{c}) a flying agent. The world is limited to 3 $\times$ 3 $\times$ 8 blocks. Each world is pictured after an empowerment maximising agent performed 1000 actions. All initial worlds are exactly like the world depicted in subfigure (\textbf{b}). Empowerment is calculated by sampling 1000 action sequences 15 steps into the future. The differences in the resulting worlds result only from the different agent embodiments (capabilities). (\textbf{a}) climbing agent; (\textbf{b}) non-climbing agent; (\textbf{c}) flying agent.}
\label{fig:embodiement}
\end{figure}

\subsubsection{Results}

Figure~\ref{fig:embodiement} shows representative resulting worlds after 1000 turns. The non-climbing agent has the most consistent behaviour, as it usually just stays where it is. Digging down would be an irreversible process because the agent would be unable to climb up again. This would thus reduce the number of reachable locations and thereby reduce the agent's empowerment. Note though that the estimated empowerment takes into account that the agent \emph{could} dig down, so it assumes that most positions below the agent are reachable. Thus, while the agent will only ever, at best, move to the nine location on the top level, it still assumes its empowerment is higher, because it could dig down, if it only wanted to Note that there exists a proposed measure of ``sustainable empowerment'' \cite{kim2009exploring} which specifically addresses this issues, by only counting those states from which the agent could return to the originating state (possibly by a \linebreak different route).

The flying agent usually digs down and then destroys the blocks. Earth blocks offer no benefit to the empowerment of the agent and are therefore just obstacles that should be removed. Keep in mind that this agent is particularly slow when digging down, as it does not fall. So, if it wants to dig down, it has to (1) take a block, (2) move down and then (3) destroy the block. Once the blocks are gone, it takes only one turn to descend. As a result, the flying agents reliably remove the blocks in the central axis. The ones on the corners of the world are often left standing (as seen in Figure~\ref{fig:fly}), as the benefit of removing them is still there but not big enough to have the agent leave its empowered position in the middle, especially once the middle axis is cleared. Since the world is limited to 8 height levels, the agent can also not simply fly up to get more space. 

The climbing agent has the highest variety in its behaviour (compared with the flying and non-climbing agent), which indicates that the different actions are very similar in empowerment, so that the random selection of sampling can affect action selection. The agent is often seen moving back and forth between two positions. The agent usually digs down first, and then uses the blocks obtained to build staircase like structures that allow it to access the upper levels of the world. The excavated structure in the bottom usually also functions as a staircase, so the agent can quickly move up and down\linebreak in the world. 

We also looked at how the estimated empowerment of an agent developed over time, and how a world restructured by an agent with one embodiment would suit an agent with a different embodiment. For typical simulations we recorded for each time-step the agent's current estimated empowerment. We also estimated the empowerment for an agent with different embodiments, if it would replace the current agent in its current position. In Figure~\ref{fig:compClimb} we can see (red line) how the climbing agent's empowerment steadily increases (if we ignore the noise of the estimation process) up to about turn 700. At this point the agent basically remains at the top of the constructed staircase. The empowerment of the non-climbing and flying agent in the final, resulting world at turn 1000 is not as high, so this restructured world offers less empowerment to them. We also see how the empowerment for the other agent varies wildly over the simulation, since the climbing agent is not sensitive to their requirements. Sometimes, the climbing agent ``accidentally'' creates a world state that is also well suited for the flying agent (for example between \linebreak turn 500 and 600), but most of the time, the other agents have less empowerment. 

The world restructured by the non-climbing agent sees little change, and consequently the graph in Figure~\ref{fig:compNoClimb} shows little movement. We see, though, that the non-climbing agent seems to be generally ill-suited to attain high empowerment. Even the world that it optimizes in regard to its own needs offers less empowerment to it than to the other agents. However, we also see that this world offers less empowerment to the flying agent and the climbing agents than the worlds they restructured to their needs. 

The graph in Figure~\ref{fig:compFly} shows how the flying agent quickly restructures the world to attain high empowerment. We again see that this world is ill-suited for the other agents. By comparing the flying and climbing agents, we can also clearly see that they both can construct worlds that are better for themselves than for the other type, so neither agent has more empowerment purely based on having a generally ``better'' embodiment. The non-climbing agent on the other hand seems to be dominated in most worlds, so its embodiment seems to be generally ill-suited to attain high empowerment.

\begin{figure}[H]

\begin{subfigure}[b]{0.3\textwidth}
\centering
\begin{tikzpicture}
\begin{axis}[ xlabel= simulation turn, ylabel = estimated reachable states , width = 15cm, height = 6cm, xmin = 0, xmax = 1000, ymin = 0, ymax = 60, grid = both, legend pos= south east
]
\addplot[red,mark size = 0.2pt] file{reachableClimb.txt};
\addplot[green,only marks,mark size = 0.2pt] file{reachableNothing.txt};
\addplot[blue,only marks,mark size = 0.2pt] file{reachableFly.txt};
\legend{climbing agent -- actual controller,non-climbing agent,flying agent}

\end{axis}
\end{tikzpicture}
\caption{}
\label{fig:compClimb}
\end{subfigure}

\begin{subfigure}[b]{0.3\textwidth}
\centering
\begin{tikzpicture}
\begin{axis}[ xlabel= simulation turn, ylabel = estimated reachable states , width = 15cm, height = 6cm, xmin = 0, xmax = 1000, ymin = 0, ymax = 60, grid = both, legend pos= south east
]
\addplot[red,only marks,mark size = 0.2pt] file{N_reachableClimb.txt};
\addplot[green,mark size = 0.2pt] file{N_reachableNothing.txt};
\addplot[blue,only marks,mark size = 0.2pt] file{N_reachableFly.txt};
\legend{climbing agent,non-climbing agent -- actual controller,flying agent}

\end{axis}
\end{tikzpicture}
\caption{}
\label{fig:compNoClimb}
\end{subfigure}

\begin{subfigure}[b]{0.3\textwidth}
\centering
\begin{tikzpicture}
\begin{axis}[ xlabel= simulation turn, ylabel = estimated reachable states , width = 15cm, height = 6cm, xmin = 0, xmax = 1000, ymin = 0, ymax = 60, grid = both, legend pos= south east
]
\addplot[red,only marks,mark size = 0.2pt] file{F_reachableClimb.txt};
\addplot[green,only marks,mark size = 0.2pt] file{F_reachableNothing.txt};
\addplot[blue,mark size = 0.2pt] file{F_reachableFly.txt};
\legend{climbing agent,non-climbing agent,flying agent -- actual controller}

\end{axis}
\end{tikzpicture}
\caption{}
\label{fig:compFly}
\end{subfigure}
\caption{Graphs showing the development of the estimated empowerment (as reachable states) for typical simulations with different agent embodiments. The lines show the empowerment for the agent that actually controls the world development, while the dots show the estimated empowerment for a given simulation turn if the actual agent is replaced by an agent with a different embodiment. (\textbf{a}) Climbing Agent; (\textbf{b}) Non-Climbing Agent;\protect\linebreak (\textbf{c}) Flying Agent.}
\label{fig:compare}
\end{figure}

\subsubsection{Discussion}

Even though all agents are controlled by the same internal motivation, we can see in Figure~\ref{fig:embodiement} that the embodiment of the agents results in quite different structures in the world. In this case, it would be possible by just looking at the resulting block configuration to determine how the agent can move.

From here on forward, we will exclusively look at the climbing agent. Not only because it displays the greatest variability, and arguably, complexity in its behaviour, but also because it became clear that the climbing agent gains most from restructuring the environment. For the flying agent, blocks are just obstacles, and the non-climbing agent just wants a flat surface, but the climbing agent can use blocks to make the world more accessible to itself. 

Embodiment also offers a good way to distinguish the effect of reaching a more empowered state in the world from restructuring the world so that it offers more empowerment. If we compare the results here with earlier examples, such as an empowered agent that moves around in a maze \cite{klyubin2005empowerment}, then we could argue that both agents just change the state of the world $R$, one by moving around, and the other by placing blocks. We model the state of the world $R$ as a random variable, and all its states are just members of an unordered set without any \emph{a priori } structure. Thus, changing the position of the agent and changing the world are both just changes to the state of $R$ that are formally of the same kind. However, embodiment gives us a clear demarcation between agent and environment, and we can ask what would happen if we remove one agent and place another somewhere in the environment? Then the second agent in the maze scenario would not profit from the first agent's actions that lead it into a better position. In the block-world case, however, a second agent could, for example, use the staircase that the first agent built, and would immediately gain a higher empowerment if it has a similar embodiment.

There is also room for further study here if we were to look at this as a case of morphological computation. Williams and Beer \cite{williams2013environmental} showed how an agent with an adapted artificial neural network produces qualitatively different behaviour based on changing embodiments and the changed associated feedback from the environment. The networks were adapted to perform a specific task with three different agent embodiments, but the evolved network had to work well in all bodies, without knowing what body it was in. The differences in behaviour resulted from the interaction of this network with the environment via the different embodiments. 
Here we have a somewhat similar case, as the control algorithm of the agents is identical, but its embodiment and the resulting, given world model, are different, resulting in different behaviour. In the future, it might also be useful to take a closer analytical look at this relationship, and possibly quantify how much exactly the different embodiments and environments contribute to the resulting world state in terms of information~\cite{zahedi2013}. 

\subsection{Experiment 2: Hazards and Planning}

We now introduce a time-varying non-adversarial (\emph{i.e.}, blind) threat to the agent, which is influenced by the setup of the environment. This is done by introducing the new element of lava. If an agent is next to a lava block (in any of the 6 directions) it will die, and subsequently all its move commands will not result in any movement. If a block below a lava block is empty, then lava will spread to that block (filling both the original and the new block). If there is an earth block below a lava block, and blocks north, south, east or west of the lava are empty, then the lava will also spread in these directions. 

\begin{figure}[htp]
\centering
\begin{subfigure}[b]{0.3\textwidth}
\begin{tikzpicture}[scale=0.6]
\cube{0}{0}{0}{20}{50}{70} 
\cube{0}{1}{0}{20}{50}{70} 
\cube{0}{2}{0}{20}{50}{70} 
\cube{0}{3}{0}{20}{50}{70} 
\cube{0}{4}{0}{20}{50}{70} 
\cube{1}{0}{0}{20}{50}{70} 
\cube{1}{1}{0}{20}{50}{70} 
\cube{1}{2}{0}{20}{50}{70} 
\cube{1}{3}{0}{20}{50}{70} 
\cube{1}{4}{0}{20}{50}{70} 
\cube{2}{0}{0}{20}{50}{70} 
\cube{2}{1}{0}{20}{50}{70} 
\cube{2}{2}{0}{20}{50}{70} 
\cube{2}{3}{0}{20}{50}{70} 
\cube{2}{4}{0}{20}{50}{70} 
\cube{3}{0}{0}{20}{50}{70} 
\cube{3}{1}{0}{20}{50}{70} 
\cube{3}{2}{0}{20}{50}{70} 
\cube{3}{3}{0}{20}{50}{70} 
\cube{3}{4}{0}{20}{50}{70} 
\cube{4}{0}{0}{20}{50}{70} 
\cube{4}{1}{0}{20}{50}{70} 
\cube{4}{2}{0}{20}{50}{70} 
\cube{4}{3}{0}{20}{50}{70} 
\cube{4}{4}{0}{20}{50}{70} 
\cube{5}{0}{0}{20}{50}{70} 
\cube{5}{1}{0}{20}{50}{70} 
\cube{5}{2}{0}{20}{50}{70} 
\cube{5}{3}{0}{20}{50}{70} 
\cube{5}{4}{0}{20}{50}{70} 
\cube{0}{0}{1}{23}{53}{73} 
\cube{0}{1}{1}{23}{53}{73} 
\cube{0}{2}{1}{23}{53}{73} 
\cube{0}{3}{1}{23}{53}{73} 
\cube{0}{4}{1}{23}{53}{73} 
\cube{1}{0}{1}{23}{53}{73} 
\cube{1}{1}{1}{23}{53}{73} 
\cube{1}{2}{1}{23}{53}{73} 
\cube{1}{3}{1}{23}{53}{73} 
\cube{1}{4}{1}{23}{53}{73} 
\cube{2}{0}{1}{23}{53}{73} 
\cube{2}{1}{1}{23}{53}{73} 
\cube{2}{2}{1}{23}{53}{73} 
\cube{2}{3}{1}{23}{53}{73} 
\cube{2}{4}{1}{23}{53}{73} 
\cube{3}{0}{1}{23}{53}{73} 
\cube{3}{1}{1}{23}{53}{73} 
\cube{3}{2}{1}{23}{53}{73} 
\cube{3}{3}{1}{23}{53}{73} 
\cube{3}{4}{1}{23}{53}{73} 
\cube{4}{0}{1}{23}{53}{73} 
\cube{4}{1}{1}{23}{53}{73} 
\cube{4}{2}{1}{23}{53}{73} 
\cube{4}{3}{1}{23}{53}{73} 
\cube{4}{4}{1}{23}{53}{73} 
\cube{5}{0}{1}{23}{53}{73} 
\cube{5}{1}{1}{23}{53}{73} 
\cube{5}{2}{1}{23}{53}{73} 
\cube{5}{3}{1}{23}{53}{73} 
\cube{5}{4}{1}{23}{53}{73} 
\cube{0}{0}{2}{26}{56}{76} 
\cube{0}{1}{2}{26}{56}{76} 
\cube{0}{2}{2}{26}{56}{76} 
\cube{0}{3}{2}{26}{56}{76} 
\cube{0}{4}{2}{26}{56}{76} 
\cube{1}{0}{2}{26}{56}{76} 
\cube{1}{1}{2}{26}{56}{76} 
\cube{1}{2}{2}{26}{56}{76} 
\cube{1}{3}{2}{26}{56}{76} 
\cube{1}{4}{2}{26}{56}{76} 
\cubeLava{2}{0}{2} 
\cubeLava{2}{1}{2} 
\cubeLava{2}{2}{2} 
\cubeLava{2}{3}{2} 
\cubeLava{2}{4}{2} 
\cube{3}{0}{2}{26}{56}{76} 
\cube{3}{1}{2}{26}{56}{76} 
\cube{3}{2}{2}{26}{56}{76} 
\cube{3}{3}{2}{26}{56}{76} 
\cube{3}{4}{2}{26}{56}{76} 
\cube{4}{0}{2}{26}{56}{76} 
\cube{4}{1}{2}{26}{56}{76} 
\cube{4}{2}{2}{26}{56}{76} 
\cube{4}{3}{2}{26}{56}{76} 
\cube{4}{4}{2}{26}{56}{76} 
\cube{5}{0}{2}{26}{56}{76} 
\cube{5}{1}{2}{26}{56}{76} 
\cube{5}{2}{2}{26}{56}{76} 
\cube{5}{3}{2}{26}{56}{76} 
\cube{5}{4}{2}{26}{56}{76} 
\cubeAgent{0}{0}{3} 
\end{tikzpicture}
\caption{}
\label{fig:beforelava}
\end{subfigure}
\begin{subfigure}[b]{0.3\textwidth}
\begin{tikzpicture}[scale=0.6]
\cube{0}{0}{0}{20}{50}{70} 
\cube{0}{1}{0}{20}{50}{70} 
\cube{0}{2}{0}{20}{50}{70} 
\cube{0}{3}{0}{20}{50}{70} 
\cube{0}{4}{0}{20}{50}{70} 
\cube{1}{0}{0}{20}{50}{70} 
\cube{1}{1}{0}{20}{50}{70} 
\cube{1}{2}{0}{20}{50}{70} 
\cube{1}{3}{0}{20}{50}{70} 
\cube{1}{4}{0}{20}{50}{70} 
\cube{2}{0}{0}{20}{50}{70} 
\cube{2}{1}{0}{20}{50}{70} 
\cube{2}{2}{0}{20}{50}{70} 
\cube{2}{3}{0}{20}{50}{70} 
\cube{2}{4}{0}{20}{50}{70} 
\cube{3}{0}{0}{20}{50}{70} 
\cube{3}{1}{0}{20}{50}{70} 
\cube{3}{2}{0}{20}{50}{70} 
\cube{3}{3}{0}{20}{50}{70} 
\cube{3}{4}{0}{20}{50}{70} 
\cube{4}{0}{0}{20}{50}{70} 
\cube{4}{1}{0}{20}{50}{70} 
\cube{4}{2}{0}{20}{50}{70} 
\cube{4}{3}{0}{20}{50}{70} 
\cube{4}{4}{0}{20}{50}{70} 
\cube{5}{0}{0}{20}{50}{70} 
\cube{5}{1}{0}{20}{50}{70} 
\cube{5}{2}{0}{20}{50}{70} 
\cube{5}{3}{0}{20}{50}{70} 
\cube{5}{4}{0}{20}{50}{70} 
\cube{0}{0}{1}{23}{53}{73} 
\cube{0}{1}{1}{23}{53}{73} 
\cube{0}{2}{1}{23}{53}{73} 
\cube{0}{3}{1}{23}{53}{73} 
\cube{0}{4}{1}{23}{53}{73} 
\cube{1}{0}{1}{23}{53}{73} 
\cube{1}{1}{1}{23}{53}{73} 
\cube{1}{2}{1}{23}{53}{73} 
\cube{1}{3}{1}{23}{53}{73} 
\cube{1}{4}{1}{23}{53}{73} 
\cube{2}{0}{1}{23}{53}{73} 
\cube{2}{1}{1}{23}{53}{73} 
\cube{2}{2}{1}{23}{53}{73} 
\cube{2}{3}{1}{23}{53}{73} 
\cube{2}{4}{1}{23}{53}{73} 
\cube{3}{0}{1}{23}{53}{73} 
\cube{3}{1}{1}{23}{53}{73} 
\cube{3}{2}{1}{23}{53}{73} 
\cube{3}{3}{1}{23}{53}{73} 
\cube{3}{4}{1}{23}{53}{73} 
\cube{4}{0}{1}{23}{53}{73} 
\cube{4}{1}{1}{23}{53}{73} 
\cube{4}{2}{1}{23}{53}{73} 
\cube{4}{3}{1}{23}{53}{73} 
\cube{4}{4}{1}{23}{53}{73} 
\cube{5}{0}{1}{23}{53}{73} 
\cube{5}{1}{1}{23}{53}{73} 
\cube{5}{2}{1}{23}{53}{73} 
\cube{5}{3}{1}{23}{53}{73} 
\cube{5}{4}{1}{23}{53}{73} 
\cube{0}{0}{2}{26}{56}{76} 
\cube{0}{2}{2}{26}{56}{76} 
\cube{0}{3}{2}{26}{56}{76} 
\cube{0}{4}{2}{26}{56}{76} 
\cube{1}{0}{2}{26}{56}{76} 
\cube{1}{1}{2}{26}{56}{76} 
\cube{1}{2}{2}{26}{56}{76} 
\cube{1}{3}{2}{26}{56}{76} 
\cube{1}{4}{2}{26}{56}{76} 
\cubeLava{2}{0}{2} 
\cubeLava{2}{1}{2} 
\cubeLava{2}{2}{2} 
\cubeLava{2}{3}{2} 
\cubeLava{2}{4}{2} 
\cube{3}{0}{2}{26}{56}{76} 
\cube{3}{1}{2}{26}{56}{76} 
\cube{3}{2}{2}{26}{56}{76} 
\cube{3}{3}{2}{26}{56}{76} 
\cube{3}{4}{2}{26}{56}{76} 
\cube{4}{0}{2}{26}{56}{76} 
\cubeAgent{4}{2}{2} 
\cube{4}{4}{2}{26}{56}{76} 
\cube{5}{0}{2}{26}{56}{76} 
\cube{5}{1}{2}{26}{56}{76} 
\cube{5}{2}{2}{26}{56}{76} 
\cube{5}{3}{2}{26}{56}{76} 
\cube{5}{4}{2}{26}{56}{76} 
\cube{2}{2}{3}{29}{59}{79} 
\end{tikzpicture}
\caption{}
\label{fig:afterlaval}
\end{subfigure}
\caption{A 5 $\times$ 6 $\times$ 5 world with a lava stream (red blocks) dividing it into two parts. The agent (blue block) starts in the smaller area, as seen in Figure~\ref{fig:beforelava}. The second graphic shows the world after 50 turns have passed, and the agent has built a ``bridge'' to access the larger part of the world. (\textbf{a}) starting configuration; (\textbf{b}) after 50 turns.}
\label{fig:lavastream}
\end{figure}
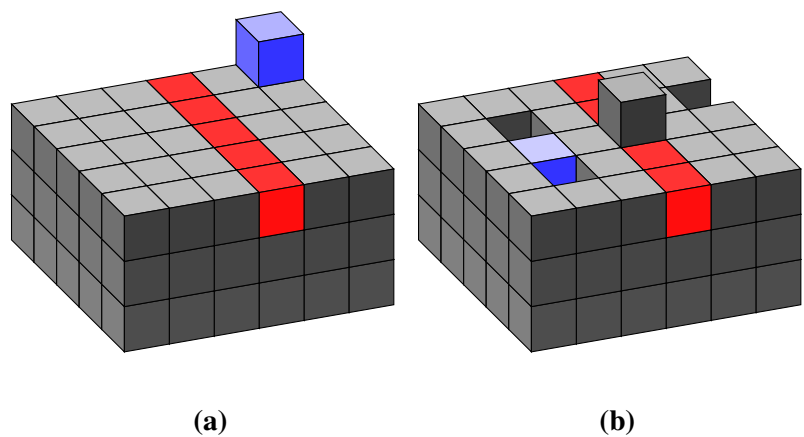

The scenario we are looking at is a 5 $\times$ 6 $\times$ 5 world, divided by a line of lava blocks, as seen in Figure~\ref{fig:beforelava}. The agent starts on the surface of the smaller area. The agent still chooses actions based on maximising its 15-step empowerment, estimated via 1000 samples. The agent does have a complete causal world model (meaning it can simulate the world ahead, given its action choices), but no further modifications to its control where made. 

\subsubsection{Results and Discussion}

Figure~\ref{fig:afterlaval} shows a representative result. The agent usually takes an earth block one block away from the lava, places it somewhere over the middle of the stream and crosses to the other side. It never touches the lava or let the lava flow into it. The agent usually keeps the lava contained, \emph{i.e.}, it does not remove blocks that allow the lava to spread. 

The first noteworthy thing is that empowerment implicitly notices the death of the agent. Death of an agent implies no effect on choosing action options, therefore vanishing empowerment. In other words, by following the empowerment-maximizing strategy, the agent implicitly avoids dying, not as a result of an external penalty related to death, but because it would lower is causal influence on the world. 

The second observation is that empowerment control seems powerful enough to string together a sequence of actions to construct a bridge, even if it only maximises empowerment for the next step. Empowerment is calculated for 15-step action sequences, but the agent only tries to get to states with high 15-step empowerment that are one step away. It does not look 15 steps ahead to find places with high empowerment. To build a bridge the agent has to pick up a block one step away from the lava stream (otherwise it would fall down and die). It then has to carry that block over, and place it on the lava. Each of those steps seems to be motivated by the local increase in empowerment. Thus, the agent picks up the block because it gives it more options, and then later places it over the lava stream, because this action now gives it more options in the current state. Observing this from the outside, it might even look as if the agent had the goal of making the other side of the world accessible to it, but this simply emerges from its internal motivation. 

It should also be noted that the ensuing behaviour becomes much more varied. After the start of the simulation, nearly all agents immediately build a bridge, but then the further actions diversify. Thus, there seems to be a much clearer gradient of empowerment leading up to this action than for the actions that follow. This is possibly because a bridge gives a large gain in empowerment, whereas most of the other structures built afterwards, such as staircases or holes in the ground, offer a lesser empowerment gain. 

\begin{figure}[H]
\centering
\begin{subfigure}[b]{0.4\textwidth}
\begin{tikzpicture}[scale=0.5]
\cube{0}{0}{0}{20}{50}{70} 
\cube{0}{1}{0}{20}{50}{70} 
\cube{0}{2}{0}{20}{50}{70} 
\cube{1}{0}{0}{20}{50}{70} 
\cube{1}{1}{0}{20}{50}{70} 
\cube{1}{2}{0}{20}{50}{70} 
\cube{2}{0}{0}{20}{50}{70} 
\cube{2}{1}{0}{20}{50}{70} 
\cube{2}{2}{0}{20}{50}{70} 
\cube{3}{0}{0}{20}{50}{70} 
\cube{3}{1}{0}{20}{50}{70} 
\cube{3}{2}{0}{20}{50}{70} 
\cube{4}{0}{0}{20}{50}{70} 
\cube{4}{1}{0}{20}{50}{70} 
\cube{4}{2}{0}{20}{50}{70} 
\cube{5}{0}{0}{20}{50}{70} 
\cube{5}{1}{0}{20}{50}{70} 
\cube{5}{2}{0}{20}{50}{70} 
\cube{6}{0}{0}{20}{50}{70} 
\cube{6}{1}{0}{20}{50}{70} 
\cube{6}{2}{0}{20}{50}{70} 
\cube{7}{0}{0}{20}{50}{70} 
\cube{7}{1}{0}{20}{50}{70} 
\cube{7}{2}{0}{20}{50}{70} 
\cube{0}{0}{1}{23}{53}{73} 
\cube{0}{1}{1}{23}{53}{73} 
\cube{0}{2}{1}{23}{53}{73} 
\cube{1}{0}{1}{23}{53}{73} 
\cube{1}{1}{1}{23}{53}{73} 
\cube{1}{2}{1}{23}{53}{73} 
\cube{2}{0}{1}{23}{53}{73} 
\cube{2}{1}{1}{23}{53}{73} 
\cube{2}{2}{1}{23}{53}{73} 
\cube{3}{0}{1}{23}{53}{73} 
\cube{3}{1}{1}{23}{53}{73} 
\cube{3}{2}{1}{23}{53}{73} 
\cube{4}{0}{1}{23}{53}{73} 
\cube{4}{1}{1}{23}{53}{73} 
\cube{4}{2}{1}{23}{53}{73} 
\cube{5}{0}{1}{23}{53}{73} 
\cube{5}{1}{1}{23}{53}{73} 
\cube{5}{2}{1}{23}{53}{73} 
\cube{6}{0}{1}{23}{53}{73} 
\cube{6}{1}{1}{23}{53}{73} 
\cube{6}{2}{1}{23}{53}{73} 
\cube{7}{0}{1}{23}{53}{73} 
\cube{7}{1}{1}{23}{53}{73} 
\cube{7}{2}{1}{23}{53}{73} 
\cube{0}{0}{2}{26}{56}{76} 
\cube{0}{1}{2}{26}{56}{76} 
\cube{0}{2}{2}{26}{56}{76} 
\cube{1}{0}{2}{26}{56}{76} 
\cube{1}{1}{2}{26}{56}{76} 
\cube{1}{2}{2}{26}{56}{76} 
\cube{2}{0}{2}{26}{56}{76} 
\cube{2}{1}{2}{26}{56}{76} 
\cube{2}{2}{2}{26}{56}{76} 
\cube{3}{0}{2}{26}{56}{76} 
\cube{3}{1}{2}{26}{56}{76} 
\cube{3}{2}{2}{26}{56}{76} 
\cube{4}{0}{2}{26}{56}{76} 
\cube{4}{1}{2}{26}{56}{76} 
\cube{4}{2}{2}{26}{56}{76} 
\cube{5}{0}{2}{26}{56}{76} 
\cube{5}{1}{2}{26}{56}{76} 
\cube{5}{2}{2}{26}{56}{76} 
\cube{6}{0}{2}{26}{56}{76} 
\cube{6}{1}{2}{26}{56}{76} 
\cube{6}{2}{2}{26}{56}{76} 
\cube{7}{0}{2}{26}{56}{76} 
\cube{7}{1}{2}{26}{56}{76} 
\cube{7}{2}{2}{26}{56}{76} 
\cube{0}{0}{3}{29}{59}{79} 
\cube{0}{1}{3}{29}{59}{79} 
\cube{0}{2}{3}{29}{59}{79} 
\cube{1}{0}{3}{29}{59}{79} 
\cube{1}{1}{3}{29}{59}{79} 
\cube{1}{2}{3}{29}{59}{79} 
\cube{2}{0}{3}{29}{59}{79} 
\cube{2}{1}{3}{29}{59}{79} 
\cube{2}{2}{3}{29}{59}{79} 
\cube{3}{0}{3}{29}{59}{79} 
\cube{3}{1}{3}{29}{59}{79} 
\cube{3}{2}{3}{29}{59}{79} 
\cube{4}{0}{3}{29}{59}{79} 
\cube{4}{1}{3}{29}{59}{79} 
\cube{4}{2}{3}{29}{59}{79} 
\cube{5}{0}{3}{29}{59}{79} 
\cube{5}{1}{3}{29}{59}{79} 
\cube{5}{2}{3}{29}{59}{79} 
\cube{6}{0}{3}{29}{59}{79} 
\cube{6}{1}{3}{29}{59}{79} 
\cube{6}{2}{3}{29}{59}{79} 
\cube{7}{0}{3}{29}{59}{79} 
\cube{7}{1}{3}{29}{59}{79} 
\cube{7}{2}{3}{29}{59}{79} 
\cubeAgent{0}{0}{4} 
\cubeLava{7}{2}{4}
\end{tikzpicture}
\caption{}
\label{fig:lava_start}
\end{subfigure}
\begin{subfigure}[b]{0.4\textwidth}
\begin{tikzpicture}[scale=0.5]
\cube{0}{0}{0}{20}{50}{70} 
\cube{0}{1}{0}{20}{50}{70} 
\cube{0}{2}{0}{20}{50}{70} 
\cube{1}{0}{0}{20}{50}{70} 
\cube{1}{1}{0}{20}{50}{70} 
\cube{1}{2}{0}{20}{50}{70} 
\cube{2}{0}{0}{20}{50}{70} 
\cube{2}{1}{0}{20}{50}{70} 
\cube{2}{2}{0}{20}{50}{70} 
\cube{3}{0}{0}{20}{50}{70} 
\cube{3}{1}{0}{20}{50}{70} 
\cube{3}{2}{0}{20}{50}{70} 
\cube{4}{0}{0}{20}{50}{70} 
\cube{4}{1}{0}{20}{50}{70} 
\cube{4}{2}{0}{20}{50}{70} 
\cube{5}{0}{0}{20}{50}{70} 
\cube{5}{1}{0}{20}{50}{70} 
\cube{5}{2}{0}{20}{50}{70} 
\cube{6}{0}{0}{20}{50}{70} 
\cube{6}{1}{0}{20}{50}{70} 
\cube{6}{2}{0}{20}{50}{70} 
\cube{7}{0}{0}{20}{50}{70} 
\cube{7}{1}{0}{20}{50}{70} 
\cube{7}{2}{0}{20}{50}{70} 
\cube{0}{0}{1}{23}{53}{73} 
\cube{0}{1}{1}{23}{53}{73} 
\cube{0}{2}{1}{23}{53}{73} 
\cube{1}{0}{1}{23}{53}{73} 
\cube{1}{1}{1}{23}{53}{73} 
\cube{1}{2}{1}{23}{53}{73} 
\cube{2}{0}{1}{23}{53}{73} 
\cube{2}{1}{1}{23}{53}{73} 
\cube{2}{2}{1}{23}{53}{73} 
\cube{3}{0}{1}{23}{53}{73} 
\cube{3}{1}{1}{23}{53}{73} 
\cube{3}{2}{1}{23}{53}{73} 
\cube{4}{0}{1}{23}{53}{73} 
\cube{4}{1}{1}{23}{53}{73} 
\cube{4}{2}{1}{23}{53}{73} 
\cube{5}{0}{1}{23}{53}{73} 
\cube{5}{1}{1}{23}{53}{73} 
\cube{5}{2}{1}{23}{53}{73} 
\cube{6}{0}{1}{23}{53}{73} 
\cube{6}{1}{1}{23}{53}{73} 
\cube{6}{2}{1}{23}{53}{73} 
\cube{7}{0}{1}{23}{53}{73} 
\cube{7}{1}{1}{23}{53}{73} 
\cube{7}{2}{1}{23}{53}{73} 
\cube{0}{0}{2}{26}{56}{76} 
\cube{0}{1}{2}{26}{56}{76} 
\cube{0}{2}{2}{26}{56}{76} 
\cube{1}{0}{2}{26}{56}{76} 
\cube{1}{1}{2}{26}{56}{76} 
\cube{1}{2}{2}{26}{56}{76} 
\cube{2}{0}{2}{26}{56}{76} 
\cube{2}{1}{2}{26}{56}{76} 
\cube{2}{2}{2}{26}{56}{76} 
\cube{3}{0}{2}{26}{56}{76} 
\cube{3}{1}{2}{26}{56}{76} 
\cube{3}{2}{2}{26}{56}{76} 
\cube{4}{0}{2}{26}{56}{76} 
\cube{4}{1}{2}{26}{56}{76} 
\cube{4}{2}{2}{26}{56}{76} 
\cube{5}{0}{2}{26}{56}{76} 
\cube{5}{1}{2}{26}{56}{76} 
\cube{5}{2}{2}{26}{56}{76} 
\cube{6}{0}{2}{26}{56}{76} 
\cube{6}{1}{2}{26}{56}{76} 
\cube{6}{2}{2}{26}{56}{76} 
\cube{7}{0}{2}{26}{56}{76} 
\cube{7}{1}{2}{26}{56}{76} 
\cube{7}{2}{2}{26}{56}{76} 
\cube{0}{0}{3}{29}{59}{79} 
\cube{0}{1}{3}{29}{59}{79} 
\cube{0}{2}{3}{29}{59}{79} 
\cube{1}{0}{3}{29}{59}{79} 
\cube{1}{1}{3}{29}{59}{79} 
\cube{1}{2}{3}{29}{59}{79} 
\cube{2}{0}{3}{29}{59}{79} 
\cube{3}{0}{3}{29}{59}{79} 
\cube{3}{2}{3}{29}{59}{79} 
\cube{4}{0}{3}{29}{59}{79} 
\cube{4}{1}{3}{29}{59}{79} 
\cube{4}{2}{3}{29}{59}{79} 
\cube{5}{0}{3}{29}{59}{79} 
\cube{5}{1}{3}{29}{59}{79} 
\cube{5}{2}{3}{29}{59}{79} 
\cube{6}{0}{3}{29}{59}{79} 
\cube{6}{1}{3}{29}{59}{79} 
\cube{6}{2}{3}{29}{59}{79} 
\cube{7}{0}{3}{29}{59}{79} 
\cube{7}{1}{3}{29}{59}{79} 
\cube{7}{2}{3}{29}{59}{79} 
\cubeLava{0}{0}{4} 
\cubeLava{0}{1}{4} 
\cubeLava{0}{2}{4} 
\cubeLava{1}{0}{4} 
\cubeLava{1}{1}{4} 
\cubeLava{1}{2}{4} 
\cubeLava{2}{0}{4} 
\cube{2}{1}{4}{32}{62}{82} 
\cube{2}{2}{4}{32}{62}{82} 
\cubeLava{3}{0}{4} 
\cube{3}{1}{4}{32}{62}{82} 
\cubeLava{3}{2}{4} 
\cubeLava{4}{0}{4} 
\cubeLava{4}{1}{4} 
\cubeLava{4}{2}{4} 
\cubeLava{5}{0}{4} 
\cubeLava{5}{1}{4} 
\cubeLava{5}{2}{4} 
\cubeLava{6}{0}{4} 
\cubeLava{6}{1}{4} 
\cubeLava{6}{2}{4} 
\cubeLava{7}{0}{4} 
\cubeLava{7}{1}{4} 
\cubeLava{7}{2}{4} 
\cubeAgent{2}{1}{5} 
\end{tikzpicture}
\caption{}
\label{fig:lava_islan}
\end{subfigure}

\begin{subfigure}[b]{0.4\textwidth}
\begin{tikzpicture}[scale=0.5]
\cube{0}{0}{0}{20}{50}{70} 
\cube{0}{1}{0}{20}{50}{70} 
\cube{0}{2}{0}{20}{50}{70} 
\cube{1}{0}{0}{20}{50}{70} 
\cube{1}{1}{0}{20}{50}{70} 
\cube{1}{2}{0}{20}{50}{70} 
\cube{2}{0}{0}{20}{50}{70} 
\cubeAgent{2}{1}{0} 
\cube{3}{1}{0}{20}{50}{70} 
\cube{3}{2}{0}{20}{50}{70} 
\cube{4}{0}{0}{20}{50}{70} 
\cube{4}{1}{0}{20}{50}{70} 
\cube{4}{2}{0}{20}{50}{70} 
\cube{5}{0}{0}{20}{50}{70} 
\cube{5}{1}{0}{20}{50}{70} 
\cube{5}{2}{0}{20}{50}{70} 
\cube{6}{0}{0}{20}{50}{70} 
\cube{6}{1}{0}{20}{50}{70} 
\cube{6}{2}{0}{20}{50}{70} 
\cube{7}{0}{0}{20}{50}{70} 
\cube{7}{1}{0}{20}{50}{70} 
\cube{7}{2}{0}{20}{50}{70} 
\cube{0}{0}{1}{23}{53}{73} 
\cube{0}{1}{1}{23}{53}{73} 
\cube{0}{2}{1}{23}{53}{73} 
\cube{1}{0}{1}{23}{53}{73} 
\cube{1}{2}{1}{23}{53}{73} 
\cube{2}{0}{1}{23}{53}{73} 
\cube{2}{2}{1}{23}{53}{73} 
\cube{3}{0}{1}{23}{53}{73} 
\cube{3}{2}{1}{23}{53}{73} 
\cube{4}{0}{1}{23}{53}{73} 
\cube{4}{2}{1}{23}{53}{73} 
\cube{5}{0}{1}{23}{53}{73} 
\cube{5}{1}{1}{23}{53}{73} 
\cube{5}{2}{1}{23}{53}{73} 
\cube{6}{0}{1}{23}{53}{73} 
\cube{6}{1}{1}{23}{53}{73} 
\cube{6}{2}{1}{23}{53}{73} 
\cube{7}{0}{1}{23}{53}{73} 
\cube{7}{1}{1}{23}{53}{73} 
\cube{7}{2}{1}{23}{53}{73} 
\cube{0}{0}{2}{26}{56}{76} 
\cube{0}{2}{2}{26}{56}{76} 
\cube{2}{1}{2}{26}{56}{76} 
\cube{2}{2}{2}{26}{56}{76} 
\cube{3}{0}{2}{26}{56}{76} 
\cube{3}{1}{2}{26}{56}{76} 
\cube{3}{2}{2}{26}{56}{76} 
\cube{4}{0}{2}{26}{56}{76} 
\cube{4}{1}{2}{26}{56}{76} 
\cube{4}{2}{2}{26}{56}{76} 
\cube{5}{0}{2}{26}{56}{76} 
\cube{5}{1}{2}{26}{56}{76} 
\cube{5}{2}{2}{26}{56}{76} 
\cube{6}{0}{2}{26}{56}{76} 
\cube{6}{1}{2}{26}{56}{76} 
\cube{6}{2}{2}{26}{56}{76} 
\cube{7}{0}{2}{26}{56}{76} 
\cube{7}{1}{2}{26}{56}{76} 
\cube{7}{2}{2}{26}{56}{76} 
\cube{0}{0}{3}{29}{59}{79} 
\cube{0}{1}{3}{29}{59}{79} 
\cube{0}{2}{3}{29}{59}{79} 
\cube{1}{0}{3}{29}{59}{79} 
\cube{1}{1}{3}{29}{59}{79} 
\cube{1}{2}{3}{29}{59}{79} 
\cube{2}{0}{3}{29}{59}{79} 
\cubeLava{2}{1}{3} 
\cubeLava{2}{2}{3} 
\cube{3}{0}{3}{29}{59}{79} 
\cube{3}{1}{3}{29}{59}{79} 
\cube{3}{2}{3}{29}{59}{79} 
\cube{4}{0}{3}{29}{59}{79} 
\cube{4}{1}{3}{29}{59}{79} 
\cube{4}{2}{3}{29}{59}{79} 
\cube{5}{0}{3}{29}{59}{79} 
\cube{5}{1}{3}{29}{59}{79} 
\cube{5}{2}{3}{29}{59}{79} 
\cube{6}{0}{3}{29}{59}{79} 
\cube{6}{1}{3}{29}{59}{79} 
\cube{6}{2}{3}{29}{59}{79} 
\cube{7}{0}{3}{29}{59}{79} 
\cube{7}{1}{3}{29}{59}{79} 
\cube{7}{2}{3}{29}{59}{79} 
\cubeLava{0}{0}{4} 
\cubeLava{0}{1}{4} 
\cube{0}{2}{4}{32}{62}{82} 
\cubeLava{1}{0}{4} 
\cubeLava{1}{1}{4} 
\cubeLava{1}{2}{4} 
\cubeLava{2}{0}{4} 
\cubeLava{2}{1}{4} 
\cubeLava{2}{2}{4} 
\cubeLava{3}{0}{4} 
\cubeLava{3}{1}{4} 
\cubeLava{3}{2}{4} 
\cubeLava{4}{0}{4} 
\cube{4}{1}{4}{32}{62}{82} 
\cubeLava{4}{2}{4} 
\cubeLava{5}{0}{4} 
\cubeLava{5}{1}{4} 
\cubeLava{5}{2}{4} 
\cubeLava{6}{0}{4} 
\cubeLava{6}{1}{4} 
\cubeLava{6}{2}{4} 
\cubeLava{7}{0}{4} 
\cubeLava{7}{1}{4} 
\cubeLava{7}{2}{4} 
\end{tikzpicture}
\caption{}
\label{fig:lava_sublevel}
\end{subfigure}
\begin{subfigure}[b]{0.4\textwidth}
\begin{tikzpicture}[scale=0.5]
\cube{0}{0}{0}{20}{50}{70} 
\cube{0}{1}{0}{20}{50}{70} 
\cube{0}{2}{0}{20}{50}{70} 
\cube{1}{0}{0}{20}{50}{70} 
\cube{1}{1}{0}{20}{50}{70} 
\cube{1}{2}{0}{20}{50}{70} 
\cube{2}{0}{0}{20}{50}{70} 
\cube{2}{1}{0}{20}{50}{70} 
\cube{2}{2}{0}{20}{50}{70} 
\cube{3}{0}{0}{20}{50}{70} 
\cube{3}{1}{0}{20}{50}{70} 
\cube{3}{2}{0}{20}{50}{70} 
\cube{4}{0}{0}{20}{50}{70} 
\cube{4}{1}{0}{20}{50}{70} 
\cube{4}{2}{0}{20}{50}{70} 
\cube{5}{0}{0}{20}{50}{70} 
\cube{5}{1}{0}{20}{50}{70} 
\cube{5}{2}{0}{20}{50}{70} 
\cube{6}{0}{0}{20}{50}{70} 
\cube{6}{1}{0}{20}{50}{70} 
\cube{6}{2}{0}{20}{50}{70} 
\cube{7}{0}{0}{20}{50}{70} 
\cube{7}{1}{0}{20}{50}{70} 
\cube{7}{2}{0}{20}{50}{70} 
\cube{0}{0}{1}{23}{53}{73} 
\cube{0}{1}{1}{23}{53}{73} 
\cube{0}{2}{1}{23}{53}{73} 
\cube{1}{0}{1}{23}{53}{73} 
\cube{1}{1}{1}{23}{53}{73} 
\cube{1}{2}{1}{23}{53}{73} 
\cube{2}{0}{1}{23}{53}{73} 
\cube{2}{1}{1}{23}{53}{73} 
\cube{2}{2}{1}{23}{53}{73} 
\cube{3}{0}{1}{23}{53}{73} 
\cube{3}{1}{1}{23}{53}{73} 
\cube{3}{2}{1}{23}{53}{73} 
\cube{4}{0}{1}{23}{53}{73} 
\cube{4}{1}{1}{23}{53}{73} 
\cube{4}{2}{1}{23}{53}{73} 
\cube{5}{0}{1}{23}{53}{73} 
\cube{5}{1}{1}{23}{53}{73} 
\cube{5}{2}{1}{23}{53}{73} 
\cube{6}{0}{1}{23}{53}{73} 
\cube{6}{1}{1}{23}{53}{73} 
\cube{6}{2}{1}{23}{53}{73} 
\cube{7}{0}{1}{23}{53}{73} 
\cube{7}{1}{1}{23}{53}{73} 
\cube{7}{2}{1}{23}{53}{73} 
\cube{0}{0}{2}{26}{56}{76} 
\cube{0}{1}{2}{26}{56}{76} 
\cube{0}{2}{2}{26}{56}{76} 
\cube{1}{0}{2}{26}{56}{76} 
\cube{1}{1}{2}{26}{56}{76} 
\cube{1}{2}{2}{26}{56}{76} 
\cube{2}{0}{2}{26}{56}{76} 
\cubeLava{2}{1}{2} 
\cube{2}{2}{2}{26}{56}{76} 
\cube{3}{0}{2}{26}{56}{76} 
\cube{3}{1}{2}{26}{56}{76} 
\cube{3}{2}{2}{26}{56}{76} 
\cube{4}{0}{2}{26}{56}{76} 
\cube{4}{1}{2}{26}{56}{76} 
\cube{4}{2}{2}{26}{56}{76} 
\cube{5}{0}{2}{26}{56}{76} 
\cube{5}{1}{2}{26}{56}{76} 
\cube{5}{2}{2}{26}{56}{76} 
\cube{6}{0}{2}{26}{56}{76} 
\cube{6}{1}{2}{26}{56}{76} 
\cube{6}{2}{2}{26}{56}{76} 
\cube{7}{0}{2}{26}{56}{76} 
\cube{7}{1}{2}{26}{56}{76} 
\cube{7}{2}{2}{26}{56}{76} 
\cube{0}{1}{3}{29}{59}{79} 
\cube{0}{2}{3}{29}{59}{79} 
\cube{1}{0}{3}{29}{59}{79} 
\cube{1}{1}{3}{29}{59}{79} 
\cube{1}{2}{3}{29}{59}{79} 
\cubeLava{2}{1}{3} 
\cubeLava{2}{2}{3} 
\cube{3}{0}{3}{29}{59}{79} 
\cube{3}{1}{3}{29}{59}{79} 
\cube{3}{2}{3}{29}{59}{79} 
\cube{4}{0}{3}{29}{59}{79} 
\cube{4}{1}{3}{29}{59}{79} 
\cube{4}{2}{3}{29}{59}{79} 
\cube{5}{0}{3}{29}{59}{79} 
\cube{5}{1}{3}{29}{59}{79} 
\cube{5}{2}{3}{29}{59}{79} 
\cube{6}{0}{3}{29}{59}{79} 
\cube{6}{1}{3}{29}{59}{79} 
\cube{6}{2}{3}{29}{59}{79} 
\cube{7}{0}{3}{29}{59}{79} 
\cube{7}{1}{3}{29}{59}{79} 
\cube{7}{2}{3}{29}{59}{79} 
\cubeAgent{1}{1}{4} 
\cube{2}{0}{4}{32}{62}{82} 
\cube{2}{1}{4}{32}{62}{82} 
\cubeLava{2}{2}{4} 
\cubeLava{3}{0}{4} 
\cubeLava{3}{1}{4} 
\cubeLava{3}{2}{4} 
\cubeLava{4}{0}{4} 
\cubeLava{4}{1}{4} 
\cubeLava{4}{2}{4} 
\cubeLava{5}{0}{4} 
\cubeLava{5}{1}{4} 
\cubeLava{5}{2}{4} 
\cubeLava{6}{0}{4} 
\cubeLava{6}{1}{4} 
\cubeLava{6}{2}{4} 
\cubeLava{7}{0}{4} 
\cubeLava{7}{1}{4} 
\cubeLava{7}{2}{4} 
\end{tikzpicture}
\caption{}
\label{fig:lava_damm}
\end{subfigure}
\caption{A 8 $\times$ 3 $\times$ 7 world scenario, where the agent (blue) starts in one corner, and a lava block (red) spreads from the other corner. The starting position is seen in Figure~\ref{fig:lava_start}. The other three images show the world after 300 turns, each is representative of a different, emerging solution. Figure~\ref{fig:lava_islan} shows an agent that constructed an island in the lava flow. Figure~\ref{fig:lava_sublevel} shows an agent that dug underground, and excavated tunnels. Figure~\ref{fig:lava_damm} shows a world where the agent managed to stop the spread of lava with a dam and trench combination. Lava does not spread, if the block below is filled with lava, so digging down one block, as seen in Figure~\ref{fig:lava_damm} in the front, makes the lava flow over the hole and then down but spread no further. (\textbf{a}) starting configuration; (\textbf{b}) Island Solution; (\textbf{c}) Cave Solution; \protect\linebreak(\textbf{d}) Dam Solution.}
\label{fig:lava_strategies}
\end{figure}

\newpage

\subsection{Experiment 3: Varying Strategies}

In this scenario we will show how the random elements of sparse sampling empowerment can lead to diverging but reasonable strategies. The initial world is depicted in Figure~\ref{fig:lava_start}. It is an 8 $\times$ 3 $\times$ 7 world, where the lower 4 levels are earth blocks. The agent starts in one corner; the other corner contains a lava block, which will spread as the scenario progresses. The lava spread speed is once every 5 turns, so the lava will only fill adjacent blocks on turns divisible by 5. The lava in the last experiment would have also spread, but was contained by blocks on the side of it. The main difference here is that the lava now starts above the ground level, so it will be initially unconstrained and will spread over the world. This experiment does not contain different starting setups or different environments; all differences result from the randomness in the agent's empowerment evaluation.

\subsubsection{Results}

\begin{figure}[H]
\begin{subfigure}[b]{1.0\textwidth}
\includegraphics[scale=1.0]{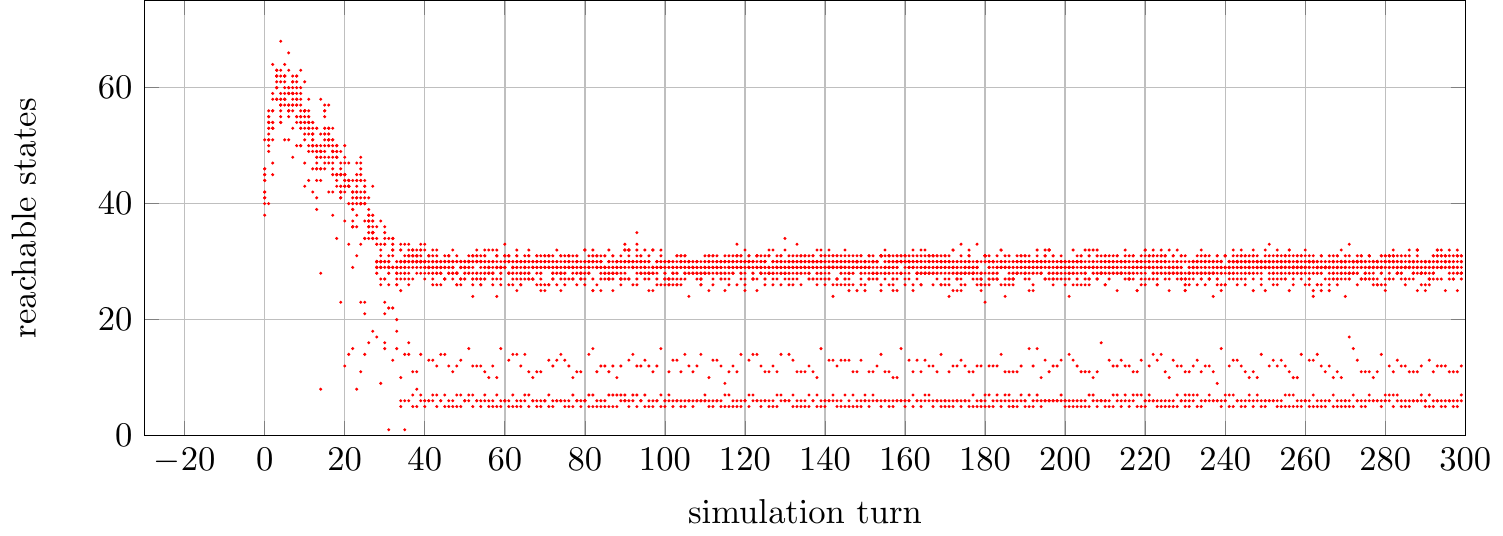}
\caption{}
\label{fig:lava_IslandE}
\end{subfigure}

\begin{subfigure}[b]{1.0\textwidth}
\includegraphics[scale=1.0]{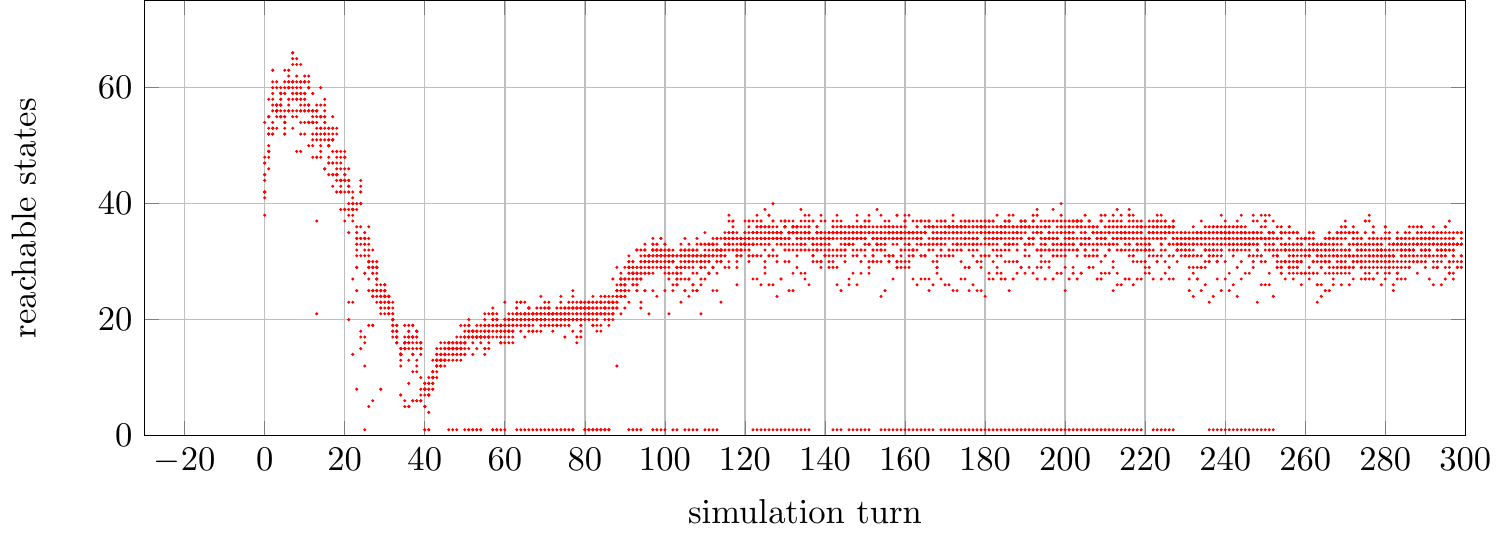}
\caption{}
\label{fig:lava_caveE}
\end{subfigure}

\begin{subfigure}[b]{1.0\textwidth}
\includegraphics[scale=1.0]{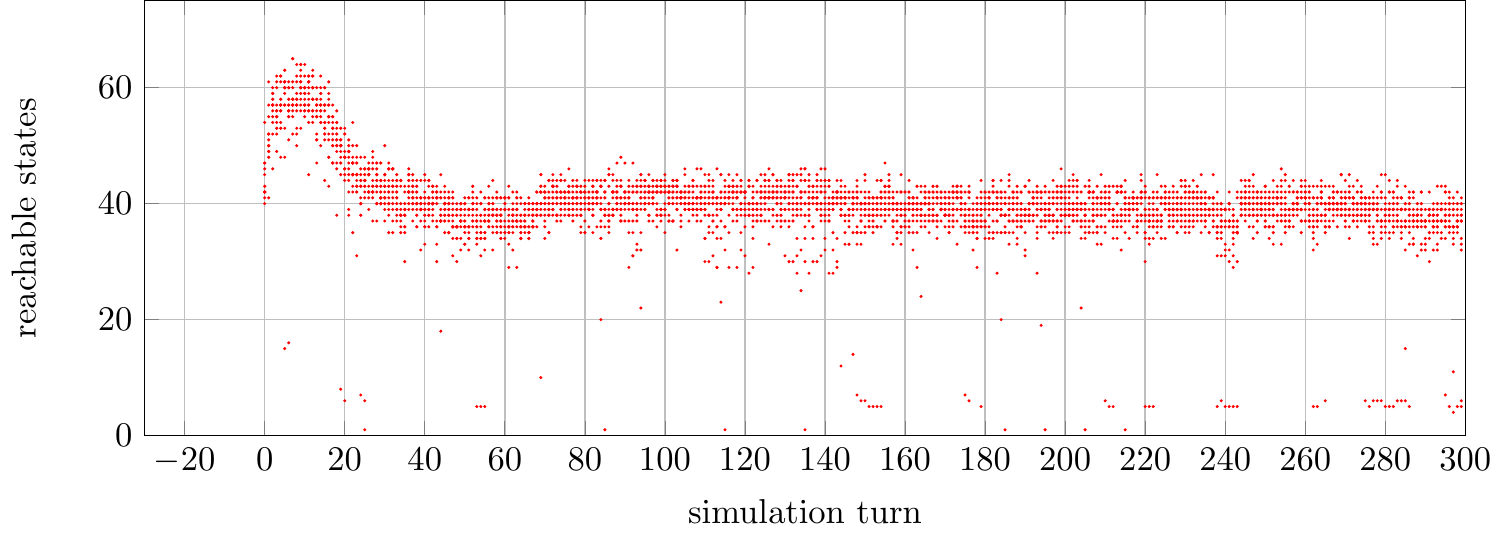}
\caption{}
\label{fig:lava_dammE}
\end{subfigure}
\caption{Three scatter plots that depict, for each turn, the different number of estimated reachable states (as an estimate for empowerment) for a given action. These are the estimates based on which the agent makes its decision, as it chooses the action having the most estimated reachable states. The three subfigures correspond to the three representative solutions depicted in Figure~\ref{fig:lava_strategies}. (\textbf{a}) Island solution; (\textbf{b}) Cave solution; (\textbf{c}) Dam solution.}
\label{fig:lava_estimated}
\end{figure}

Observing multiple runs, we could see that the agent initially moved to the middle of the world and performed a variety of different actions, \emph{i.e.}, moving around, digging, placing blocks. The initial spread of the lava only cuts off very few locations, so the increasing hazard of the whole world being covered with lava does not influence the agent behaviour much. This is also illustrated by the plots in Figure~\ref{fig:lava_estimated}, which show the estimated number of states for the different actions for each turn, upon which the agent makes its decision. 
All three graphs, depicting trials with the same starting configuration, begin in a similar fashion. The agent moves to the middle, and we see that the empowerment for the different possible actions (more precisely the empowerment of the state this action leads to) rises because the agents are all moving towards the middle of the block. We can also see that the estimated number of reachable states are very similar (all between 65 and 55), which means that randomness in the sampling could easily change their ordering. 

All the graphs in Figure~\ref{fig:lava_estimated} then show a drop in empowerment values starting around turn 10. This is the effect of the lava starting to constrain the agent's movement. As the lava starts to cover more and more of the world the agents start to react to the lava, as more and more of the sampled action sequences are cut short, and the empowerment loss of the agent becomes more noticeable to the estimator. 

Depending on how the world has been shaped up to this point, the agent then usually focuses on one of the three strategies depicted in Figure~\ref{fig:lava_strategies}. In about 50 observed trials no agent ever died, \emph{i.e.}, came in contact with the lava. 

The most common strategy when the lava starts to cover the world is for the agent to put down one block on the lava level, and step onto this little island. If there are already some other blocks on the level of the lava, the agent might then try to connect them with little bridges to regain some mobility. However, in most cases, the resulting situation looks like Figure~\ref{fig:lava_islan}, where the agent constructed a small island. Sometimes this island is just one block large, and the agent does nothing. The development of the empowerment values for this can be seen in Figure~\ref{fig:lava_IslandE}. After the island is constructed around turn 40, we see that the evaluation for the different actions separates into distinct groups. There are now actions that will lead to the death of an agent in the imminent future (like stepping off the island), but they are not chosen, as their empowerment value is clearly lower than the value for other options.

The second most common strategy for the agent is to go underground. Here the agent digs down a hole, and covers it with a block on top. The agent then digs further down, and excavates an underground network of tunnels in which it can move around. The agent also takes care not to puncture the ceiling, so no lava can flow into the underground cave. The empowerment development for this strategy can be seen in Figure~\ref{fig:lava_caveE}. Once the agent drops into the cave (around turn 40) the empowerment value is momentarily very low, for all possible actions. Nevertheless, it then starts to excavate the underground level and regains a lot of empowerment. In the end, the agent living underground has more options and mobility than the agent living in an island, which is reflected by the higher empowerment value in the graph. Also note that there are occasional levels of 0 after turn 40; these correspond to the turns in which the agent is right under the ceiling, and taking a block would result in the lava flowing on top of the agent and killing it. Since the estimated empowerment of this option is very low, it is not chosen. Keep in mind that the points in Figure~\ref{fig:lava_estimated} denote the estimated empowerment of the possible actions, so having one low point means the agent is currently in a position where it could take actions that would severely reduce the empowerment in the resulting state. 

Putting up a series of connected blocks or removing a series of connected blocks can create a dam or trench that stops the lava. If the agent has already, mostly incidentally, placed some blocks earlier, it can then remove or place some additional blocks to stop the lava from spreading. An example of this can be seen in Figure~\ref{fig:lava_damm}, where the agent retains a small surface area. The two blocks in the front act as a dam, while the lava on the third row is blocked by a hole in the ground. Again, Figure~\ref{fig:lava_dammE} shows the empowerment development. We can basically see that the construction of the dam and trench around turn 30 stops the spread of the lava and stabilizes the agent's empowerment level. Occasional low empowerment values correspond to actions that would damage the dam and make the remaining area overflow with lava. 

\subsubsection{Discussion}

This scenario demonstrates how the agents will adapt their strategy to the current state of the environment once the lava situation becomes pressing. The agent acts in a way that seems sensible, from an outside perspective, where it manages to avoid getting killed by lava in all trials. This scenario also shows the two main features of sparse compared with regular empowerment. The initial variability in the behaviour is due to the randomness of the sampling process. The effect of the randomness is more powerful in the initial phase, because the estimated empowerment values for several actions are very similar to each other, so the variable introduced by the random sampling is more likely to permute their preference ordering. The threat of the lava is, initially, far removed, so it has little effect on the agent's decision making. Once the lava is closer to the agent the differences in the empowerment of possible actions become larger, so the random sampling has less of an influence on the agent's behaviour, and the agents will reliably take those actions that will keep them alive. 

We would like to point out that the main interest lies in how this behaviour is generated and not in the fact that it can be generated at all. It is safe to assume that a reinforcement learning approach that would apply a large negative reward to touching lava would also produce an agent that would avoid lava. However, such an agent would only be concerned with the lava, while the empowerment-based agent also proceeds to excavate a cave to improve its mobility, or might build a bridge to the disconnected islands on top of the lava. It would of course be possible to explicitly specify additional rewards to create similar behaviour, but this behaviour would be less impressive, if all aspects of it had to be separately motivated by an external reward, rather than being task-independent. This distinction between external and internal motivation becomes more problematic if we look at this as a model of natural evolved agents. It is possible to adapt internalized specific reward structures, such as the avoidance of pain or extreme heat, which allow the agents to deal with specific external circumstances. In the end, the avoidance of pain is similarly intrinsic to the motivation to be empowered, and both are capable of producing goal-directed behaviour, because the goals derive from the fulfilment of this internal motivation. A real organic agent would much more likely avoid the lava because of the incurred pain rather than based on some thought of limited future options. However, what we wanted to demonstrate here is the versatility of empowerment---the ability to generate reasonable behaviour for a range of situations. Ideally, we want an intrinsic motivation that would work in new situations where innate, long evolved reflexes or instincts (or hand-designed algorithms, in the case of artificial agents) would fail. For instance, in an environment that changes more rapidly than reflexes can be trained, it seems easier to evolve a few generic mechanisms that produce a range of behaviours, rather than to produce a separate behaviour to build bridges, a separate behaviour to build stairs, \emph{etc.}

\section{Future Work}

While the initial results of using sparsely sampled empowerment seem promising for defining ways for agents to restructure the world, there are several directions in which this work could be extended, to both make the computation faster and increase the applicability to different scenarios.

To deal with events that are further into the future, it would also be helpful to extend the time \linebreak horizon of the agent. With the classical, exhaustive empowerment algorithm, this quickly becomes too computationally expensive. By using a fixed number of samples, we can ensure that the computation time only grows linearly, but since longer time horizons usually lead to more possible outcomes, the empowerment estimation becomes worse (as seen in Section \ref{sec:sparse}). This could be alleviated by using more sophisticated methods, such as impoverished empowerment \cite{anthony2011impoverished,anthony13:_gener_self_motiv_strat_ident}. Impoverished empowerment evaluates all action sequences for a shorter time horizon and then chooses those that contribute the most to the empowerment as new action primitives. Also, Monte-Carlo tree search \cite{browne2012survey} methods have recently seen a lot of development, and have proven useful to focus exploration on those branches of the tree that contribute to the agent's desired utility function. It might be possible to adapt similar techniques to empowerment calculations. 

We have previously discussed that the selection of only parts of the sensor input for the empowerment calculation can focus the resulting control on different aspects of the simulation \cite{salge2014book}. In our examples, the resulting sensor input of the agent was reduced to its position, focusing the agent's control on mobility, and on structures that would increase its mobility. Previous empowerment related work \cite{capdepuy2007maximization} on multi-agent simulations equipped agents with density sensors that could measure the number of other agents around them. This led to the emergence of agent clustering structures, as agents needed to stay together, to generate different sensor input states. Similarly, the agent's sensors in the block world simulations could be extended to register the different blocks directly around them, which could generate incentives to build a variety of different structures. 

Another obvious extension is the study of several agents that jointly restructure an environment. Previous studies of empowerment in agent collectives \cite{capdepuy2012perception,capdepuy2007maximization} demonstrated that coordinated agents are able to improve their overall empowerment. However, more recent work showed that not jointly coordinated and essentially unpredictable agents can also provide a source of noise, which lowers the empowerment of an agent interacting with them \cite{salge2013empowerment}. Eventually, it comes down to how one models an agent's view of other agents, specifically in regard to how well one agent can predict another agent's actions, and how much other agent's follow cues or orders from one another. While agents could still ``cooperate'' by just improving the world in different parts far away from each other, there are indications that a closer interactive cooperation needed for jointly constructing structures would require some degree of attention, communication and prediction modelling of the other agent. Furthermore, given that other agents are just one form of noise in the environment, it would also be good to extend the current scenarios to include different forms of noise. Empowerment, in general, deals well with noise, and the original formalism requires no modification. Unfortunately, the computation of channel capacity for a noisy channel is much more complicated than for discrete and deterministic channels \cite{blahut1972computation,arimoto1972}. Furthermore, in the case of noise, the sparse sampling approach is also not directly feasible as a computation reduction tool, so the horizon length and actions that we are dealing with in the block world scenario are too computationally expensive for now. However, if it is possible to create similar estimations for empowerment for non-deterministic block worlds, then we could possibly observe a whole range of new phenomena. 

\section{Conclusions}

We investigated the possibility of utilizing a simple and straightforward sampling estimate for\linebreak empowerment in a deterministic world and studied how well it would perform in an \linebreak environment-changing scenario. The results in this paper indicate that empowerment provides a sufficient intrinsic motivation for an agent to manipulate and restructure its environment in a way that is both reflective of its own embodiment and reactive to changing environmental conditions. This includes the generation of actions that ensure the agent's survival in threatening situations. As they are indications that the fundamental principles behind empowerment may hold for more generic domains, such as noisy and continuous worlds, we expect it to be possible, once the technical problems of computability are addressed, to apply empowerment-motivated control to a range of artificial agents or robots. 

There have been robotic applications \cite{leu2013} where empowerment has been combined with a classical AI path-planning method to enhance robotic following behaviour. In light of the current paper's results, it would be interesting to look at a robot that can change its environment or its own embodiment within it, such as the robot presented in \cite{brodbeck2012robotic}, which could use hot meld adhesives to modify its effector. In the original experiment, the robot's aim was to optimize its performance in an externally specified water scooping task. It would be interesting to see whether empowerment could provide an intrinsic motivation for the robot to self-modify, aiming not specifically for scooping water efficiently but just to have more causal influence on the world, given its current embodiment and the environment. 

Coming back to our original starting point of ``Minecraft'', there is also a promising avenue of application to computer games here, namely the generation of more ``natural'' behaviour for non-player entities. Currently, NPC behaviour is usually hard-coded to perform actions leading to specific goals that were assigned by the designer. With a good formalism for intrinsic motivation, NPC behaviour could be generated with goals that are meaningful for the NPCs because they are based on their own digital affordances \cite{gibson1979ecological} and their own experiential world. While technically challenging, this could provide far greater flexibility in NPC reactions to a changing world, which could address the problem of providing meaningful reactions to the wide and unpredictable behaviour of players in open world games. 

In summary, empowerment fulfils the requirements we earlier set out for an intrinsic motivation and generates actions that are meaningfully grounded in an agent's embodiment. We believe that empowerment can act as part of an ensemble of complementing intrinsic motivations. In turn, the aspects of intrinsic motivation that make it essential to the developments of humans\cite{oudeyer2007intrinsic} are likely just as necessary and essential for the development of artificial intelligence that could engage us and the world on eye level with human intelligence. 

\acknowledgements{Acknowledgments}

This research was supported by the European Commission as part of the CORBYS (Cognitive Control Framework for Robotic Systems) project under contract FP7 ICT-270219 ({www.corbys.eu). The views expressed in this paper are those of the authors and not necessarily those of the consortium.

\section*{\noindent Conflicts of Interest}
\vspace{12pt}

The authors declare no conflict of interest.


\end{document}